%% file: main.tex
\documentclass[11pt]{article}

\usepackage[preprint]{acl}

\usepackage{times}
\usepackage{latexsym}

\usepackage[T1]{fontenc}

\usepackage[utf8]{inputenc}

\usepackage{microtype}

\usepackage{inconsolata}

\usepackage{graphicx}

\usepackage{booktabs}
\usepackage{amsmath}
\usepackage{amssymb}
\usepackage[table]{xcolor}
\usepackage{algorithm}
\usepackage{algorithmic}

\usepackage{url}

\title{OPUS: A Simple yet Effective Unified Framework \\ for Open-Vocabulary Detection}

\author{
  \textbf{Xiaoyan Wei\textsuperscript{1}},
  \textbf{Zhimin Yao\textsuperscript{1}},
  \textbf{Ruilin Yang\textsuperscript{1}},
  \textbf{Wei Zhang\textsuperscript{1}},
  \textbf{Yong Dai\textsuperscript{2}},
  \textbf{Yi Zhang\textsuperscript{2}},
  \textbf{Wei Ge\textsuperscript{1}\footnotemark[2]}
  \\
  \\
  \textsuperscript{1} Megvii Technology Inc.
  \textsuperscript{2} X-Humanoid
  \\
  \small{
    \textbf{Correspondence:} \href{gewei@megvii.com}{gewei@megvii.com}
  }
}

\begin{document}
\maketitle

\footnotetext[2]{Corresponding author.}

\input{sections/0_abstract}
\input{sections/1_introduction}
\input{sections/2_relatedwork}
\input{sections/3_method}
\input{sections/4_experiments}
\input{sections/5_conclusion}


\section*{Limitations}

Although OPUS achieves strong overall performance, generic visual prompting remains less improved than text and interactive visual prompting.
Our training uses Visual-I-style prompts sampled from ground-truth instances, whereas Visual-G inference relies on cross-image category prototypes aggregated from reference images.
This training-inference mismatch makes Visual-G more sensitive to reference quality and prototype noise, especially for rare categories.
Our analysis suggests that current generic visual prompting still requires more targeted modeling, such as better reference image selection, prototype aggregation, and prototype denoising.

In addition, our SAM3-based data engine is designed to scale heterogeneous grounding supervision in a single pass, but its effectiveness depends on the quality and domain coverage of the underlying promptable model.
We validate OPUS on standard natural-image detection benchmarks, including COCO, LVIS-minival, and ODinW35, but have not evaluated the annotation pipeline on specialized domains such as medical, industrial, or remote-sensing imagery.
Extending and validating the pipeline in such domains remains future work.



\bibliography{main}
\clearpage
\appendix

\input{sections/6_appendix}

\end{document}

%% file: sections/0_abstract.tex
\begin{abstract}

Recent unified open-vocabulary detection (OVD) supports heterogeneous prompts, including text queries, visual exemplars, and their combinations, but often rely on increasingly complex designs such as heavy cross-modal fusion, staged training, and iterative annotation pipelines.
We revisit whether such complexity is necessary in the era of stronger foundation models.
Our finding is that unified OVD can be made substantially simpler with semantic-rich visual representations and scalable grounding supervision.
We present OPUS (\textbf{O}pen-vocabulary, \textbf{P}rompt-\textbf{U}nified, \textbf{S}imple), a unified detector supporting text, interactive visual, generic visual, and mixed prompting within one framework.
OPUS adopts a simple three-part design. Its model architecture combines a semantic-rich visual encoder, built on a DINOv3-ConvNeXt-B backbone with efficient hybrid encoding, with a prompt-aware decoder that avoids prompt-specific branches for unified prompt reasoning. OPUS is trained with a one-stage text-visual training strategy with Instance-level Contrastive Alignment (ICA), and is supported by a SAM3-based single-pass data engine for heterogeneous grounding supervision.
Experiments on COCO, LVIS-minival, and ODinW35 show that OPUS achieves state-of-the-art Visual-I performance, reaching 68.1/69.2/54.7 AP, while maintaining balanced Text and Visual-G accuracy.
OPUS also turns mixed prompting from interference into complementarity, improving over text or visual prompt alone.
These results show that simplicity and strong unified prompting capability can be achieved together.

\end{abstract}

%% file: sections/1_introduction.tex
\section{Introduction}
\label{sec:introduction}

\begin{figure*}[t]
    \centering
    \includegraphics[width=\textwidth]{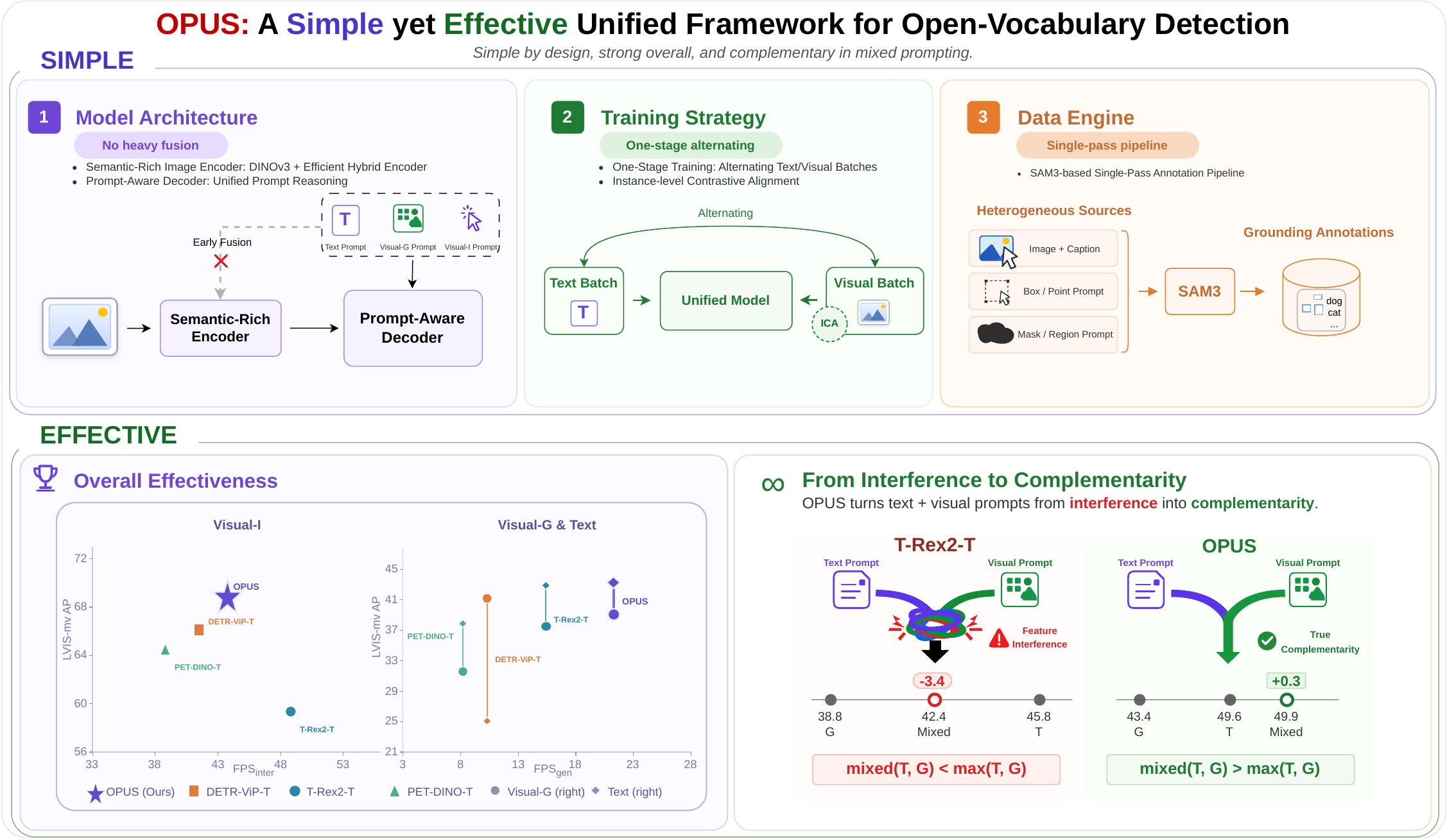}
    \caption{
    Overview of OPUS.
    Top: OPUS simplifies unified open-vocabulary detection through a DINOv3-based semantic-rich encoder with a prompt-aware decoder, one-stage text-visual training with Instance-level Contrastive Alignment (ICA), and a SAM3-based single-pass data engine.
    Bottom left: OPUS achieves a strong accuracy-efficiency trade-off across prompting settings.
    Bottom right: OPUS turns mixed prompting from interference into complementarity, improving over text or visual prompt alone.
    }
    \label{fig:framework}
\end{figure*}

Open-vocabulary detection (OVD) aims to localize and recognize objects beyond a fixed label set by aligning visual regions with semantic embeddings.
Driven by vision-language pretraining and promptable architectures, OVD has evolved from text-prompted detection toward unified prompting, where detectors also accept visual prompts such as reference images, boxes, or points.
Compared with text prompts, visual prompts provide complementary appearance-level cues and enable finer control in long-tailed or textually ambiguous scenarios.

However, this flexibility has often been achieved through increasingly complex system designs.
Architecturally, recent unified OVD methods~\citep{CP-DETR,Detr-ViP}, built upon Grounding DINO~\citep{GroundingDINO}, rely on heavy early cross-modal fusion and additional decoder-side prompt interaction modules. 
In training strategy, T-Rex2~\citep{T-Rex2} uses staged text-visual alternating training, while PET-DINO~\citep{PET-DINO} further introduces more prompt-enriched training schemes.
In data construction, methods such as T-Rex2~\citep{T-Rex2} depend on iterative re-annotation pipelines that interleave model training and data refinement.
These advances substantially improve unified OVD.
At the same time, strong results are often specialized to particular prompt modalities or system designs, making it unclear how to achieve more balanced performance across text and visual prompting.

Modern foundation models suggest a simpler alternative.
Strong visual encoders such as DINOv3~\citep{Dinov3} provide semantic-rich and transferable image representations, while promptable segmentation models such as SAM3~\citep{SAM3} enable scalable grounding annotation without iterative relabeling.
Together, these advances motivate us to revisit whether heavy cross-modal fusion, staged training, and iterative relabeling are necessary for unified OVD.
\textit{Our results suggest that they are not: with strong visual representations and scalable single-pass grounding supervision, unified OVD can be made substantially simpler.}

We instantiate this principle as OPUS (\textbf{O}pen-vocabulary, \textbf{P}rompt-\textbf{U}nified, \textbf{S}imple), a simple unified OVD framework supporting text, interactive visual, generic visual, and mixed prompting.
OPUS adopts a simple three-part design:
(1) a model architecture that combines a semantic-rich visual encoder, built on a DINOv3-ConvNeXt-B backbone with efficient hybrid encoding, with a prompt-aware decoder that avoids prompt-specific branches for unified prompt reasoning;
(2) a one-stage text-visual training strategy with an Instance-level Contrastive Alignment (ICA) objective for fine-grained instance coordination;
and (3) a SAM3-based single-pass data pipeline that converts text-image, classification, and segmentation sources into unified grounding supervision.

Despite its simple design, OPUS achieves state-of-the-art Visual-I performance on COCO~\citep{COCO}, LVIS-minival~\citep{LVIS}, and ODinW35~\citep{ODinW}, while maintaining competitive Text and Visual-G performance.
On Visual-I of ODinW35 , OPUS outperforms PET-DINO-T~\citep{PET-DINO} by +6.4 AP and DETR-ViP-T~\citep{Detr-ViP} by +7.9 AP.

More importantly, OPUS achieves genuine mixed-prompt complementarity: combining Text and Visual-G prompts improves performance from 49.6 to 49.9 on COCO and from 43.0 to 45.2 on LVIS-minival, whereas T-Rex2 drops from 45.8 to 42.4 on COCO.
Further analysis reveals a \textit{role shift} in how the two modalities contribute to mixed prompting:
as grounding supervision expands, the Mixed-over-Text gain decreases from 8.7 to 2.2, while the Mixed-over-Visual-G gain increases from 1.6 to 6.6.
This suggests that Visual-G becomes less effective as a standalone prompt and increasingly relies on textual semantics in the mixed setting, making stronger Visual-G modeling a promising future direction.

Our contributions are summarized as follows:
\begin{itemize}
\item We present OPUS, a simple unified OVD framework that supports text, visual, and mixed prompting without heavy cross-modal fusion, staged training, or iterative relabeling.

\item Experiments on COCO, LVIS-minival, and ODinW35 show that OPUS achieves state-of-the-art Visual-I performance while maintaining strong Text and Visual-G accuracy within a single unified model.

\item We demonstrate genuine mixed-prompt complementarity and reveal an evolving role shift between text and visual prompts, offering new insights into unified prompt interaction.
\end{itemize}

%% file: sections/2_relatedwork.tex
\section{Related Work}
\label{sec:relatedwork}

\paragraph{From Text-Prompted Detection to Visual Prompting.}
Open-vocabulary detection was first dominated by text-prompted paradigms, where category names or noun phrases serve as detection queries.
Representative methods such as OWL-ViT~\citep{OWL-ViT}, GLIP~\citep{GLIP}, and Grounding DINO~\citep{GroundingDINO} establish this paradigm through large-scale vision-language pretraining and region-text alignment, while later approaches including OV-DINO~\citep{OV-DINO}, DetCLIPv3~\citep{DetCLIPv3}, LLMDet~\citep{LLMDet}, and YOLO-World~\citep{YOLO-World} further improve open-world generalization and inference efficiency.
OpenSeeD~\citep{OpenSeed} studies joint open-vocabulary segmentation and detection under text prompts.
However, text prompts provide limited appearance-level guidance, especially for fine-grained, long-tailed, or textually ambiguous categories.
Visual prompting methods such as DINOv~\citep{DINOv}, T-Rex~\citep{T-Rex}, MQDet~\citep{MQDet}, CountGD~\citep{CountGD}, and Prompt-DINO~\citep{Prompt-DINO} address this limitation by using exemplars, boxes, points, or multimodal queries to provide instance-level appearance cues.
These works show that visual prompts are complementary to text, but they often focus on specific prompting settings rather than a fully unified text-visual prompting framework.

\paragraph{Unified Prompting and System Complexity.}
Recent work has therefore moved toward unified detectors that support both text and visual prompts.
T-Rex2~\citep{T-Rex2} introduces a unified text-visual prompting paradigm through staged text-to-visual optimization and alternating training.
YOLOE~\citep{YOLOE} extends the YOLO-style detection framework to support unified text and visual prompting through a lightweight design, providing a real-time and cost-efficient solution.
PET-DINO~\citep{PET-DINO} further adopts prompt-enriched multi-stage optimization built upon Grounding DINO~\citep{GroundingDINO}.
In parallel, methods such as CP-DETR~\citep{CP-DETR} and DETR-ViP~\citep{Detr-ViP} strengthen prompt interaction through additional encoder or decoder-side cross-modal fusion modules.
Beyond model designs and training strategies, recent works also scale grounding supervision in different ways.
Some expand image-text pairs via detector-generated pseudo boxes, as in GLIP~\citep{GLIP}, OWLv2~\citep{OWLv2}, and YOLO-World~\citep{YOLO-World}. Others build generative annotation engines with vision-language models, such as DetCLIPv3~\citep{DetCLIPv3} and WeDetect~\citep{WeDetect}. Still others assemble large multi-source corpora, including GLEE~\citep{GLEE} and DINO-X~\citep{DINO-X}, while T-Rex2~\citep{T-Rex2} designs separate text and visual data engines with iterative relabeling. 
These approaches broaden open-vocabulary coverage, yet collecting and cleaning such data remain costly and often hard to reproduce.

\paragraph{Foundation Models for Simpler Unified OVD.}
Modern foundation models offer a different route.
Strong self-supervised visual encoders such as DINOv3~\citep{Dinov3} provide transferable and semantically rich image representations, reducing the need for heavy early cross-modal fusion.
Meanwhile, promptable segmentation models such as SAM3~\citep{SAM3} make it possible to generate large-scale grounding supervision from heterogeneous sources without iterative relabeling.
Rather than further increasing architectural, optimization, or data-engine complexity, our work revisits unified OVD under these stronger foundation-model priors.
We show that a simple framework can support text, interactive visual, generic visual, and mixed prompting while maintaining strong performance across prompting settings.

%% file: sections/3_method.tex
\section{Method}
\label{sec:method}


OPUS is a unified open-vocabulary detector designed around three simple components: a semantic-rich image encoder with a prompt-aware decoder, a one-stage text-visual training strategy, and a single-pass grounding data engine.
Given an image and a set of heterogeneous prompts, OPUS maps text and visual prompts into a shared prompt space and decodes object queries conditioned on both image features and prompt features.
Figure~\ref{fig:framework} provides an overview of our framework.

\subsection{Model Architecture}


\paragraph{Semantic-Rich Image Encoder.}
We use a DINOv3~\citep{Dinov3} pretrained ConvNeXt~\citep{ConvNeXt} backbone to extract multi-scale image features $\{C_3, C_4, C_5\}$.
Instead of performing heavy early cross-modal fusion, OPUS first processes these visual features with an efficient hybrid encoder~\citep{RT-DETR}, which consists of attention-based intra-scale feature interaction and CNN-based cross-scale feature fusion:
\begin{equation}
    \mathbf{F} = \mathrm{HybridEnc}(C_3, C_4, C_5) \in \mathbb{R}^{N_f \times D},
\end{equation}
where $N_f = \sum_{l \in \{3,4,5\}} H_l W_l$ denotes the number of visual tokens across all feature scales, and $D$ is the hidden dimension.
The strong semantic transferability of DINOv3 allows OPUS to rely on this lightweight visual encoding pipeline without encoder-side image-text fusion.

\paragraph{Unified Prompt Generator.}
Following T-Rex2~\citep{T-Rex2}, 
OPUS supports text and visual prompts under a unified prompt representation.
For text prompts, category names or noun phrases are encoded by a CLIP~\citep{CLIP} text encoder and projected into the detector hidden space, producing text prompt features $\mathbf{P}^t \in \mathbb{R}^{M \times D}$, where $M$ is the number of category prompts in the current input.
For visual prompts, a Visual Prompt Encoder (VPE) encodes user-provided boxes or points.
Given $K_m$ visual exemplars for category $m \in \{1,\ldots,M\}$, VPE stacks three Transformer layers with multi-scale deformable cross-attention to extract features from multi-scale image feature maps conditioned on these exemplars, yielding instance-level content prompt features $\mathbf{P}^v_{m,{\mathrm{con}}}$ and an aggregated class-level prompt feature $\mathbf{P}^v_{m,{\mathrm{cls}}}$ via a learnable class token:
\begin{equation}
    \mathbf{P}^v_m = [\mathbf{P}^v_{m,{\mathrm{con}}};\mathbf{P}^v_{m,{\mathrm{cls}}}]
    \in \mathbb{R}^{(K_m+1)\times D}.
\end{equation}

Class-level prompt features of all $M$ categories are then stacked into $\mathbf{P}^v_{\mathrm{cls}}$:
\begin{equation}
\mathbf{P}^v_{\mathrm{cls}} = [\mathbf{P}^v_{1,\mathrm{cls}};\mathbf{P}^v_{2,\mathrm{cls}};\cdots;\mathbf{P}^v_{M,\mathrm{cls}}] \in \mathbb{R}^{M \times D}.
\end{equation}

Depending on the prompting setting, the final prompt representation $\mathbf{P}\in \mathbb{R}^{M \times D}$ can be constructed from textual prompts, visual prompts, or their combination:
\begin{equation}
\mathbf{P} =
\begin{cases}
\mathbf{P}^t, & \text{Text}, \\
\mathbf{P}^v_{\mathrm{cls}}, & \text{Visual}, \\
(\mathbf{P}^t+\mathbf{P}^v_{\mathrm{cls}})/2, & \text{Mixed}.
\end{cases}
\end{equation}

\paragraph{Prompt-Aware Decoder.}
To support unified reasoning over heterogeneous prompts, we adopt a prompt-aware decoder based on the Grounding DINO's~\citep{GroundingDINO} decoder.
Given image features $\mathbf{F} \in \mathbb{R}^{N_f \times D}$ and prompt features $\mathbf{P} \in \mathbb{R}^{M \times D}$, we initialize $N_q$ object queries $\mathbf{Q} \in \mathbb{R}^{N_q \times D}$ through prompt-guided proposal selection.
Each decoder layer $l$ iteratively refines the queries via query self-attention, prompt cross-attention over $\mathbf{P}$, and image cross-attention over $\mathbf{F}$:
\begin{equation}
    \mathbf{Q}^{(l)} =
    \mathrm{DecLayer}\!\left(\mathbf{Q}^{(l-1)}, \mathbf{F}, \mathbf{P}\right).
    \label{eq:decoder}
\end{equation}
The resulting decoder naturally supports text, interactive visual, generic visual, and mixed prompting within a unified decoding process, without introducing prompt-specific architectural branches.

\subsection{Training Strategy and Objective}


\paragraph{One-Stage Text-Visual Training.}
OPUS is trained with a one-stage alternating strategy.
For text-prompt batches, we use object detection and visual grounding data, where category names or noun phrases serve as text prompts.
For visual-prompt batches, we sample ground-truth boxes or points as visual prompts and train the model to detect objects conditioned on visual exemplars.
Text and visual batches are alternated with a fixed ratio during training.

\paragraph{Detection and Alignment Losses.}
The detection objective follows DETR-style detection training and includes classification, box regression, IoU, and denoising losses:
\begin{equation}
    \mathcal{L}_{\mathrm{det}} =
    \mathcal{L}_{\mathrm{cls}}
    + \mathcal{L}_{\mathrm{bbox}}
    + \mathcal{L}_{\mathrm{iou}}
    + \mathcal{L}_{\mathrm{dn}}.
\end{equation}
For visual-prompt batches, we additionally use the global prompt alignment loss from T-Rex2~\citep{T-Rex2}, which aligns each class-level visual prompt $\mathbf{P}^v_{\mathrm{cls}}$ with its matched text prompt $\mathbf{P}^t$.
The base objective is therefore
\begin{equation}
    \mathcal{L}_{\mathrm{base}} =
    \mathcal{L}_{\mathrm{det}}
    + \mathcal{L}_{\mathrm{align}}.
\end{equation}
While this global alignment encourages text-visual category consistency, it does not explicitly supervise instance-level visual cues, which are important for visual and mixed prompting.

\paragraph{Instance-level Contrastive Alignment (ICA).}
To strengthen instance-level coordination between visual prompts and object queries, we introduce ICA objective.
Let $\{(Q_i, P^v_{\mathrm{con},i})\}_{i=1}^{N^+}$ denote the query-content prompt pairs associated through Hungarian matching between predicted queries and ground-truth instances.
ICA treats each matched pair $\left(Q_i,P^v_{\mathrm{con},i}\right)$ as positive and the remaining content prompts in the same mini-batch as negatives.
The ICA loss is then defined as:
\begin{equation}
\mathcal{L}_{\mathrm{ICA}}
=
-\frac{1}{N^+}
\sum_{i=1}^{N^+}
\log
\frac{
\exp(\mathrm{sim}(Q_i, P^v_{\mathrm{con},i})/\tau)
}{
\sum\limits_{j=1}^{N^+}
\exp(\mathrm{sim}(Q_i, P^v_{\mathrm{con},j})/\tau)
},
\end{equation}
where $\mathrm{sim}(\cdot,\cdot)$ denotes cosine similarity and $\tau$ is a fixed temperature hyperparameter.
The final objective is
\begin{equation}
    \mathcal{L}
    =
    \mathcal{L}_{\mathrm{base}}
    + \mathcal{L}_{\mathrm{ICA}},
\end{equation}
where $\mathcal{L}_{\mathrm{ICA}}$ is also applied only to visual-prompt batches and set to zero otherwise.
ICA encourages visual prompt features to preserve fine-grained instance information and improves their coordination with object queries.

\begin{figure}[t]
    \centering
    \includegraphics[width=\columnwidth]{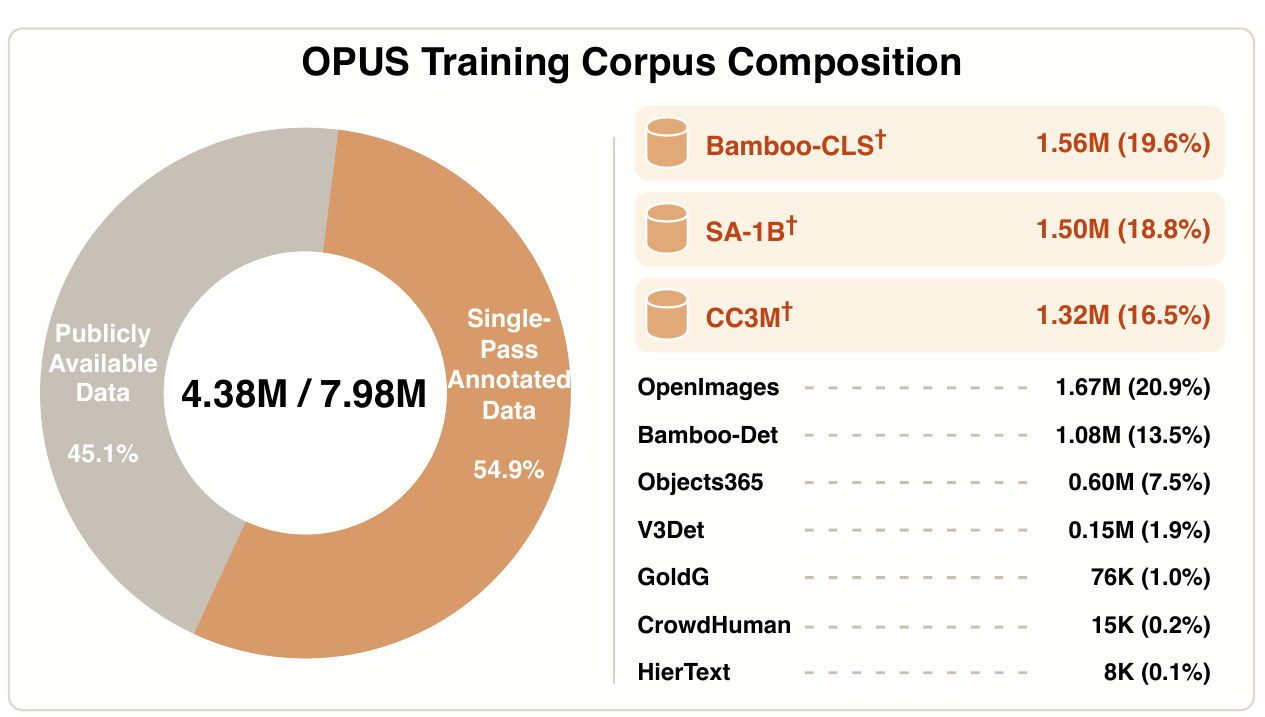}
    \caption{
    Composition of the OPUS training corpus.
    OPUS uses 7.98M training images from heterogeneous detection, grounding, classification, image-text, and segmentation sources.
    Datasets marked with $^\dagger$ are generated by our SAM3-based single-pass annotation pipeline: Bamboo-CLS, CC3M, and SA-1B.
    Together, these generated annotations contribute 4.38M images, accounting for 54.9\% of the corpus, showing that OPUS scales grounding supervision without iterative relabeling.
    }
    \label{fig:data_engine}
\end{figure}





\subsection{Data Engine}

To construct heterogeneous grounding supervision, we build a SAM3-based single-pass annotation pipeline.
The training corpus combines existing detection and grounding datasets with single-pass annotated data from text-image, classification, and segmentation sources, as summarized in Figure~\ref{fig:data_engine}.

For text-image data, we extract noun phrases from captions and use them as text prompts for SAM3~\citep{SAM3} to generate corresponding object regions.
For classification data, we use image-level category labels as prompts and convert the resulting SAM3 regions into box annotations.
For segmentation data, we treat existing masks or regions as geometric prompts to recover additional object instances and convert them into bounding boxes, and apply TAP~\citep{TAP} to assign unified semantic labels to the resulting regions.

This pipeline produces Bamboo-CLS~\citep{Bamboo}, CC3M~\citep{CC3M}, and annotated SA-1B~\citep{SA-1B} supervision for text and visual prompting.
To control annotation noise, we apply a unified post-processing pipeline, including confidence thresholding, category-wise and image-wise NMS, removal of near full-image boxes, and filtering of oversized crowd boxes.
Process details and representative success and failure cases are provided in Appendix~\ref{appendix:data-engine}.

%% file: sections/4_experiments.tex
\section{Experiments}
\label{sec:experiments}

\input{tables/main_results}

\subsection{Experimental Setup}

\paragraph{Model and Training Details.}
Unless otherwise specified, OPUS refers our final setting with a DINOv3 pretrained ConvNeXt-B backbone, a CLIP-B text encoder, and the Visual Prompt Encoder (VPE) following T-Rex2~\citep{T-Rex2}.
The hidden dimension of all FFN layers is 1024.
We optimize with AdamW~\citep{adamW}, using a learning rate of $4 \times 10^{-5}$ for the image backbone and text encoder, and $4 \times 10^{-4}$ for the remaining modules. 
Models are trained on 16 A100 GPUs with a global batch size of 128. Vsual-prompt and text-prompt batches are alternated at a ratio of 1:8. 
For the O365~\citep{O365}+GoldG~\citep{GoldG} ablation, we train on 8 A100 GPUs with a batch size of 64 for 120k iterations.
Full hyperparameters are provided in Appendix~\ref{appendix:training_details}.

\paragraph{Evaluation Protocol.}
Following T-Rex2, we evaluate zero-shot detection on COCO~\citep{COCO}, LVIS-minival~\citep{LVIS}, and ODinW35~\citep{ODinW}, reporting standard AP metrics as in prior work. 
AP numbers in comparison tables are taken from the original papers.
We consider four prompting settings.
\textit{Text} uses category names directly as text prompts.
\textit{Visual-G} constructs category-level visual prototypes by randomly sampling $N{=}16$ reference images per category from the training set.
\textit{Visual-I} simulates interactive detection by randomly sampling one ground-truth instance per category from each test image as the visual prompt.
\textit{Mixed} averages Text and Visual-G embeddings.
For efficiency, we report $\mathrm{FPS}_{\mathrm{gen}}$ for generic detection and $\mathrm{FPS}_{\mathrm{inter}}$ for interactive detection on COCO, using our reproductions of publicly available codebases or model architectures from prior work.
Measurement details are provided in Appendix~\ref{appendix:eval_details}.

\subsection{Main Results}

\paragraph{Accuracy and Efficiency.}
Table~\ref{tab:main_results} compares OPUS with T-Rex2-T, CP-DETR-T, DETR-ViP-T, PET-DINO-T, and YOLOE-v8-L under Visual-I, Visual-G, and Text prompting.
OPUS (Ours) achieves the best Visual-I performance among these baselines on COCO, LVIS-minival, and ODinW35, exceeding DETR-ViP-T by 2.7 on COCO and 3.1 AP on LVIS-minival and surpassing PET-DINO-T and DETR-ViP-T on ODinW35 by 6.4 and 7.9 AP, respectively.
Prior methods often specialize in one prompt mode (e.g., DETR-ViP-T has strong Visual-G but weak Text), whereas OPUS remains more balanced across prompting modes.

For a controlled comparison, we also report OPUS (Swin-T) under the same Swin-T~\citep{Swin} backbone and 800$\times$1333 input as T-Rex2-T.
It improves Visual-I LVIS-minival by 11.3 AP over T-Rex-T, with only a minor Visual-G trade-off.
OPUS (Ours) is our final practical setting with a DINOv3 pretrained ConvNeXt-B backbone, hybrid encoding and a reduced 640$\times$640 input.
It further recovers Visual-G relative to OPUS (Swin-T) and yields a more balanced profile across generic and interactive detection.
Despite the lower resolution, it remains competitive in accuracy while running faster than the Swin-T baselines in Table~\ref{tab:main_results}, achieving the highest $\mathrm{FPS}_{\mathrm{gen}}$ of 21.3.

As a cross-scale reference, Table~\ref{tab:visual_i_large} further compares Visual-I against larger Swin-L models.
Although DETR-ViP-L remains stronger on COCO and LVIS-minival, OPUS (Ours) stays within about 3 AP of it on these two benchmarks and leads ODinW35 by 3.5 AP, despite using a smaller ConvNeXt-B backbone and lower input resolution.

\input{tables/compare_to_large_models}

\paragraph{Mixed Prompting Complementarity.}
\input{tables/mix_prompt_results}
Table~\ref{tab:mixed} evaluates Mixed prompting that combines Text and Visual-G at inference.
On COCO, T-Rex2-T drops from 45.8 (Text) to 42.4 (Mixed), showing interference.
In contrast, OPUS improves from 49.6 to 49.9 on COCO and from 43.0 to 45.2 on LVIS-minival.
On full LVIS, T-Rex2-T achieves stronger Visual-G than OPUS, yet its Mixed score remains lower.
These results suggest that OPUS does not simply rely on stronger individual prompt modalities, but more effectively combines text semantics and visual appearance cues.
Section~\ref{sec:analysis} further analyzes how this complementarity evolves as grounding supervision expands.

\input{tables/ablation_component}

\subsection{OPUS Design Analysis}

\paragraph{Component-wise Performance.}
\label{ablation:componentwise_analysis}

We study how each design step contributes to the final OPUS configuration.
Since T-Rex2 has not released its code or model weights, we re-implement its architecture based on the published paper and train it on O365 and GoldG as our \textbf{baseline (step~A)}.
Table~\ref{tab:component_wise_ablation} reports the progression on LVIS-minival as we add each OPUS component.

\textbf{Semantic-rich enc.}
Replacing Swin-T with a DINOv3-based ConvNeXt-B backbone, reducing the input from 800$\times$1333 to 640$\times$640, and adopting a hybrid encoder improves all prompting modes and raises $\mathrm{FPS}_{\mathrm{gen}}$ from 15.4 to 22.0 (+43\%).

\textbf{Prompt-aware dec.}
Adding the prompt-aware decoder improves Visual-I from 57.5 to 63.0 and Text from 29.5 to 32.7, but reduces Visual-G from 35.3 to 33.3.
Visual-I already knows which categories are present in the image, so prompt and image cross-attention helps find similar instances. Text also remains effective under supervised training with both positive and negative category names.
In contrast, Visual-G uses cross-image category prototypes at inference, which can mismatch the Visual-I-style prompts seen during training.
This mismatch is most severe for rare categories, whose prototypes are aggregated from fewer and more diverse instances.

\textbf{ICA.}
Table~\ref{tab:ica_ablation} isolates ICA under O365+GoldG after Step~C.
Visual prompting is effective on the long tail.
Even without ICA, Visual-G already outperforms Text on rare categories (AP$_r$ 26.1 vs.\ 20.5).
Adding ICA raises Visual-G AP$_r$ to 30.0 (Text AP$_r$ remains 20.3), while overall Visual-I, Visual-G, and Text AP change by less than 0.5.
Mixed AP also increases by 1.5 (36.5$\rightarrow$38.0).
Section~\ref{sec:analysis} further analyzes why ICA helps rare-category Visual-G beyond class-level alignment $\mathcal{L}_{\mathrm{align}}$.

\textbf{Data engine.}
ICA and the data engine address complementary long-tailed bottlenecks.
ICA strengthens long-tailed visual prompting, while the data engine expands grounding supervision and mainly improves text semantic coverage, lifting Text from 33.2 to 43.0 and Mixed from 38.0 to 45.2 on LVIS-minival, with Text AP$_r$ rising from 20.3 to 36.6.

\paragraph{Component-wise Efficiency.}
Table~\ref{tab:component_wise_efficiency} decomposes step~B into B$_1$ to B$_3$ (C to E follow Table~\ref{tab:component_wise_ablation}).
Switching to DINOv3-ConvNeXt-B at 800$\times$1333 (B$_1$) increases Params/GFLOPs and slows inference (FPS 15.4$\rightarrow$11.6).
Reducing the input to 640$\times$640 (B$_2$) cuts GFLOPs from 439 to 208 and recovers FPS to 19.4.
The hybrid encoder (B$_3$) further reduces GFLOPs to 164 and reaches FPS 22.0.
Later steps add little cost.
The prompt-aware decoder adds 2M parameters and 1 GFLOPs, while ICA and the data engine leave inference unchanged.
Appendix~\ref{appendix:efficiency} repoprts the same IDs with per-step accuracy under all prompting modes.

\input{tables/ablation_ica}
\input{tables/ablation_efficiency}

\subsection{Grounding Supervision and Data Scaling}
\input{tables/ablation_data_scaling}

\paragraph{Data Engine Annotation Quality.}

The SAM3-based data engine contributes 4.38M of the 7.98M training images (54.9\%), so annotation quality matters for the scaling results below.
We randomly sample 200 images from each of Bamboo-CLS, CC3M, and SA-1B and manually check every pseudo box--phrase pair after filtering.
A box is marked incorrect if either its localization or label is wrong.
Box accuracies are 91.9\%, 89.5\% and 91.2\% on the three sources (Table~\ref{tab:data_engine_quality}).
Residual errors are mainly nested or redundant background boxes on Bamboo-CLS and CC3M, and incorrect box--label assignments on SA-1B.
Appendix~\ref{appendix:data-engine} details the filtering pipeline and shows qualitative success and failure cases.

\paragraph{Progressive Data Scaling.}

Table~\ref{tab:data_scaling} first reports progressive data scaling on T-Rex2-T (re-impl.), with official T-Rex2-T as a reference.
Each row lists all datasets used in that run.
The impact is much stronger on LVIS-minival than on COCO.
On COCO, Visual-I, Visual-G, and Text change only sightly from O365+GoldG to the largest set, moving from 57.2, 41.3, and 45.8 to 56.0, 41.4, and 47.1.
On LVIS-minival, Visual-G rises from 33.6 to 40.5 and Text from 26.5 to 45.2, indicating that large-scale grounding mainly improves long-tailed coverage.
Visual-I is already strong with O365 visual training and further improves on LVIS-minival after adding SA-1B.
Visual-G stays stable on COCO and improves on LVIS-minival, unlike official T-Rex2, whose Visual-G drops after adding visual data (Appendix~\ref{appendix:staged_trex2}).

\paragraph{Data Scaling Across Frameworks.}

Table~\ref{tab:data_scaling} compares T-Rex2-T (re-impl.), OPUS (Swin-T) and OPUS (Ours) under the same baseline of O365+GoldG and the same largest training set.
On LVIS-minival Visual-I, scaling to the full set improves T-Rex2-T (re-impl.) by 3.1 AP, while OPUS improves by 6.5 AP with Swin-T and by 6.0 AP with ConvNeXt-B. On the full set, OPUS (Swin-T) and OPUS (Ours) reach 70.6 and 69.2 Visual-I AP on LVIS-minival and 68.1 on COCO, compared with 59.4 and 56.0 for T-Rex2-T (re-impl.). 
Visual-G shows a different scaling pattern within OPUS. 
OPUS (Swin-T) gains only 1.4 AP on LVIS-minival and slightly drops on COCO, whereas OPUS (Ours) gains 5.1 AP on LVIS-minival. On the full set, OPUS (Ours) leads COCO Visual-G at 43.4, while T-Rex2-T (re-impl.) remains higher on LVIS-minival at 40.5. 
For Text, OPUS (Ours) is strongest on COCO at 49.6, while T-Rex2-T (re-impl.) remains slightly higher on LVIS-minival at 45.2. 
Overall, OPUS converts expanded grounding data more effectively into interactive visual prompting gains, and data scaling works best together with the full OPUS design for more balanced improvements across prompting modes.

\subsection{Further Analysis}
\label{sec:analysis}

\paragraph{Fine-grained Category Analysis Across Component Evolution.}
\begin{figure}[!t]
\centering
\includegraphics[width=\linewidth]{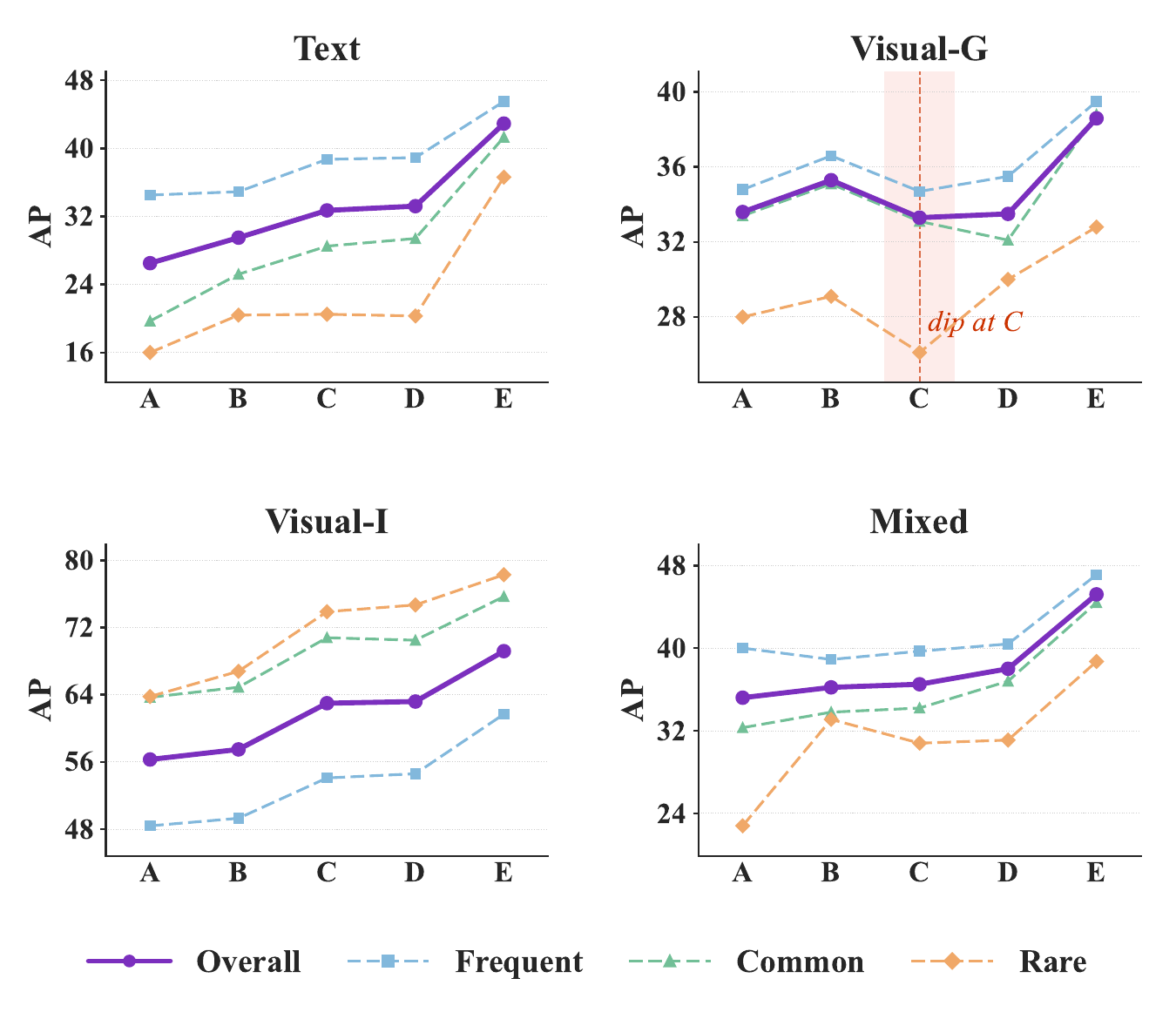}
\caption{
Fine-grained category analysis across component evolution on LVIS-minival.
We report overall AP and AP on frequent, common, and rare categories under Text, Visual-G, Visual-I, and Mixed prompting.
}
\label{fig:fcr_ablation}
\end{figure}

Figure~\ref{fig:fcr_ablation} breaks down the A$\to$E component evolution into frequent, common, and rare categories on LVIS-minival.
Overall AP and category-group AP mostly improve throughout the progression, confirming that the gains are not limited to a specific frequency group.

The main exception appears when adding the prompt-aware decoder.
While this decoder improves Text and Visual-I, it temporarily hurts Visual-G, especially on rare categories.
As discussed in Table~\ref{tab:component_wise_ablation}, this likely stems from a training-inference mismatch: the decoder is trained with Visual-I-style prompts from the same image, whereas Visual-G inference uses cross-image category prototypes.
Rare categories are more sensitive to this mismatch because their prototypes are aggregated from fewer and more diverse instances.

ICA partially recovers this drop, with the most notable improvement on rare categories.
This suggests that the issue is not only the distribution mismatch of Visual-G prototypes, but also the limited granularity of class-level prompt alignment.
The base alignment objective $\mathcal{L}_\text{align}$ pulls the aggregated visual prototype $\mathbf{P}^v_\text{cls}$ toward the text prompt feature $\mathbf{P}^t$.
While useful for category-level alignment, this objective may suppress fine-grained visual details, especially for rare categories whose text representations and visual prototypes are both less reliable.
By contrast, ICA aligns instance-level visual prompt features with their matched decoder queries.
This provides a finer-grained learning signal that is less dependent on class-level text anchors, helping recover rare-category Visual-G performance.

\paragraph{Evolving Complementarity Between Text and Visual Prompting.}
Table~\ref{tab:component_wise_ablation} also reveals how text-visual complementarity evolves across the A$\to$E progression.
Mixed prompting consistently outperforms both Text and Visual-G inference, showing that the two prompt types provide complementary cues.
However, the source of this complementarity changes over time: $\Delta_T$ decreases from 8.7 to 2.2, while $\Delta_G$ increases from 1.6 to 6.6.

This trend reflects \textit{a progressive role shift between the two prompting modalities}.
At early stages, Text prompting suffers from limited long-tail semantic coverage, so Visual-G provides useful appearance-level guidance.
As the data engine expands grounding supervision, Text prompting becomes much stronger and eventually surpasses Visual-G.
Mixed prompting then shifts from using visual prompts to compensate for weak text semantics toward using text prompts to stabilize noisier cross-image visual prototypes.
This evolving interaction explains why mixed prompting remains consistently beneficial, and also exposes a limitation of current generic visual prompting, suggesting that the Visual-G paradigm itself deserves further study, including reference selection, prototype aggregation, and prototype denoising.

%% file: tables/main_results.tex
\begin{table*}[!t]
\centering
\small
\renewcommand{\arraystretch}{1.5}
\setlength{\tabcolsep}{5pt}
\definecolor{vi}{HTML}{F0F0FA}
\definecolor{gt}{HTML}{F0F7F0}
\definecolor{mb}{HTML}{F5F5F5}

\resizebox{\textwidth}{!}{%
\begin{tabular}{
  >{\columncolor{mb}}l
  >{\columncolor{mb}}l
  >{\columncolor{mb}}c
  >{\columncolor{vi}}c >{\columncolor{vi}}c >{\columncolor{vi}}c >{\columncolor{vi}}c
  >{\columncolor{gt}}c >{\columncolor{gt}}c >{\columncolor{gt}}c
  >{\columncolor{gt}}c >{\columncolor{gt}}c >{\columncolor{gt}}c
  >{\columncolor{gt}}c
}
\hline

\cellcolor{mb} & \cellcolor{mb} & \cellcolor{mb}
& \multicolumn{4}{c}{\cellcolor{vi}\textbf{Visual-I}}
& \multicolumn{3}{c}{\cellcolor{gt}\textbf{Visual-G}}
& \multicolumn{3}{c}{\cellcolor{gt}\textbf{Text}}
& \cellcolor{gt} \\
\hline

\cellcolor{mb}\textbf{Model}
& \cellcolor{mb}\textbf{Backbone}
& \cellcolor{mb}\textbf{Input Size}

& \cellcolor{vi}\textbf{COCO}
& \cellcolor{vi}\textbf{LVIS-mv}
& \cellcolor{vi}\textbf{ODinW}
& \cellcolor{vi}\textbf{FPS\textsubscript{inter}}

& \cellcolor{gt}\textbf{COCO}
& \cellcolor{gt}\textbf{LVIS-mv}
& \cellcolor{gt}\textbf{ODinW}

& \cellcolor{gt}\textbf{COCO}
& \cellcolor{gt}\textbf{LVIS-mv}
& \cellcolor{gt}\textbf{ODinW}

& \cellcolor{gt}\textbf{FPS\textsubscript{gen}} \\
\hline

T-Rex2-T
& Swin-T
& 800$\times$1333

& 56.6 & 59.3 & 37.7 & \textbf{\textit{48.8}}

& 38.8 & 37.4 & 23.6

& \multicolumn{1}{|>{\columncolor{gt}}c}{45.8}
& 42.8 & 18.0

& \underline{\textit{15.4}} \\

CP-DETR-T
& Swin-T
& 800$\times$1333

& 61.8 & 64.1 & 41.0 & --

& -- & -- & --

& \multicolumn{1}{|>{\columncolor{gt}}c}{\textbf{52.0}}
& \textbf{47.6} & \textbf{27.3}

& -- \\

DETR-ViP-T
& Swin-T
& 800$\times$1333

& 65.4 & 66.1 & 46.8 & \textit{41.5}

& \underline{43.2} & \textbf{41.1} & \textbf{31.2}

& \multicolumn{1}{|>{\columncolor{gt}}c}{--}
& $^\dagger$25.0 & --

& \textit{10.3} \\

PET-DINO-T
& Swin-T
& 800$\times$1333

& 64.3 & 64.5 & 48.3 & \textit{38.8}

& 38.4 & 31.5 & 25.5

& \multicolumn{1}{|>{\columncolor{gt}}c}{\underline{49.8}}
& 37.8 & 20.6

& \textit{8.2} \\

YOLOE-v8-L
& YOLOv8-L
& 640$\times$640

& -- & -- & -- & --

& -- & 34.2 & --

& \multicolumn{1}{|>{\columncolor{gt}}c}{--}
& 35.9 & -- & -- \\

\hline

\textbf{OPUS (Swin-T)}
& Swin-T
& 800$\times$1333

& \underline{68.1} & \textbf{70.6} & \underline{53.7} & \textit{42.1}

& 35.8 & 35.8 & 24.8

& \multicolumn{1}{|>{\columncolor{gt}}c}{47.1} 
& \underline{44.6} & 22.0

& \textit{14.5} \\

\textbf{OPUS (Ours)}
& ConvNeXt-B
& 640$\times$640

& \textbf{68.1} & \underline{69.2} & \textbf{54.7} & \underline{\textit{43.5}}

& \textbf{43.4} & \underline{38.6} & \underline{25.9}

& \multicolumn{1}{|>{\columncolor{gt}}c}{49.6}
& 43.0 & \underline{22.1}

& \textbf{\textit{21.3}} \\
\hline

\end{tabular}%
}
\caption{
Main zero-shot detection results under Visual-I, Visual-G, and Text prompting.
AP is reported on COCO, LVIS-minival (LVIS-mv), and ODinW35, abbreviated as ODinW in the table.
$\mathrm{FPS}_{\mathrm{inter}}$ measures Visual-I inference with cached image features, and $\mathrm{FPS}_{\mathrm{gen}}$ measures Text/Visual-G inference with cached prompt features.
OPUS achieves the best Visual-I performance while maintaining balanced accuracy across prompting modes.
\textbf{Bold}: best; \underline{underline}: second best.
$^\dagger$: result from supplementary materials.
}
\label{tab:main_results}
\end{table*}

%% file: tables/compare_to_large_models.tex
\begin{table}[!t]
\centering
\small
\renewcommand{\arraystretch}{1.1}
\setlength{\tabcolsep}{6pt}
\resizebox{\columnwidth}{!}{%
\begin{tabular}{llccc}
\hline

\textbf{Model}
& \textbf{Backbone}
& \textbf{COCO}
& \textbf{LVIS-mv}
& \textbf{ODinW} \\

\hline

T-Rex2-L
& Swin-L
& 58.5
& 62.5
& 39.7 \\

CP-DETR-L
& Swin-L
& \textcolor{gray}{68.4}
& \textcolor{gray}{71.6}
& 50.6 \\

DETR-ViP-L
& Swin-L
& \textbf{71.1}
& \textbf{71.9}
& \underline{51.2} \\

PET-DINO-L
& Swin-L
& 66.5
& 65.8
& 49.7 \\

\hline

\textbf{OPUS(Ours)}
& ConvNeXt-B
& \underline{68.1}
& \underline{69.2}
& \textbf{54.7} \\

\hline
\end{tabular}%
}

\caption{
Visual-I comparison against Swin-L baselines.
\textcolor{gray}{Gray} numbers indicate settings where the model used the corresponding benchmark's training split during training or pre-training; these numbers are reported for reference but excluded from ranking.
\textbf{Bold} and \underline{underline} mark the best and second best among ranked results.
}
\label{tab:visual_i_large}
\end{table}

%% file: tables/mix_prompt_results.tex
\begin{table}[!t]
\centering
\setlength{\tabcolsep}{2.5pt}
\renewcommand{\arraystretch}{1.08}
\resizebox{\columnwidth}{!}{%
\begin{tabular}{@{}lccc@{}}
\hline
\textbf{Model} 
& \textbf{COCO} 
& \textbf{LVIS-mv} 
& \textbf{LVIS} \\
\textbf{} 
& Text/Visual-G/Mixed
& Text/Visual-G/Mixed
& Text/Visual-G/Mixed \\
\hline
T-Rex2-T 
& 45.8/38.8/42.4 
& 42.8/37.4/-- 
& \textbf{34.8}/\textbf{34.9}/37.0 \\
\hline
\textbf{OPUS(Ours)} 
& \textbf{49.6/43.4/49.9} 
& \textbf{43.0/38.6/45.2} 
& \textbf{34.8}/31.2/\textbf{37.6} \\
\hline
\end{tabular}%
}
\caption{
Mixed prompting comparison on COCO, LVIS-minival (LVIS-mv), and LVIS.
\textbf{Bold} indicates the best or tied best.
$-$ indicates that the result is not reported.
}
\label{tab:mixed}
\end{table}

%% file: tables/ablation_component.tex
\begin{table*}[!t]
\centering
\small
\setlength{\tabcolsep}{6pt}
\renewcommand{\arraystretch}{1.1}
\begin{tabular}{clcccc|cc|c}
\hline
\textbf{ID} & \textbf{Evolution Step}
& \textbf{Visual-I}
& \textbf{Visual-G}
& \textbf{Text}
& \textbf{Mixed}
& $\mathbf{\Delta_T}$
& $\mathbf{\Delta_G}$
& \textbf{FPS\textsubscript{gen}} \\
\hline

A & Baseline (Swin-T)
& 56.3 & 33.6 & 26.5 & 35.2
& +8.7 & +1.6
& 15.4 \\

B & + Semantic-rich enc.
& 57.5 & 35.3 & 29.5 & 36.2
& +6.7 & +0.9
& 22.0 \\

C & + Prompt-aware dec.
& 63.0 & 33.3 & 32.7 & 36.5
& +3.8 & +3.2
& 21.3 \\

D & + ICA
& 63.2 & 33.5 & 33.2 & 38.0
& +4.8 & +4.5
& 21.3 \\

E & + Data engine
& \textbf{69.2}
& \textbf{38.6}
& \textbf{43.0}
& \textbf{45.2}
& +2.2 & +6.6
& 21.3 \\
\hline
\end{tabular}

\caption{
Component-wise ablation and prompt-complementarity analysis on LVIS-minival.
Mixed denotes Text+Visual-G prompting.
$\Delta_T$ and $\Delta_G$ measure the gains of Mixed prompting over Text-only and Visual-G-only prompting, respectively.
}
\label{tab:component_wise_ablation}
\end{table*}

%% file: tables/ablation_ica.tex
\begin{table}[!t]
\centering
\footnotesize
\renewcommand{\arraystretch}{1.1}
\setlength{\tabcolsep}{6pt}
\resizebox{\columnwidth}{!}{%
\begin{tabular}{lcccccccc}
\hline
\textbf{ICA}
& \multicolumn{2}{c}{\textbf{Visual-I}}
& \multicolumn{2}{c}{\textbf{Visual-G}}
& \multicolumn{2}{c}{\textbf{Text}}
& \multicolumn{2}{c}{\textbf{Mixed}} \\
\cline{2-9}
& \textbf{AP} & \textbf{AP$_r$}
& \textbf{AP} & \textbf{AP$_r$}
& \textbf{AP} & \textbf{AP$_r$}
& \textbf{AP} & \textbf{AP$_r$} \\
\hline
w/o
& 63.0 & 73.9 & 33.3 & 26.1 & 32.7 & 20.5 & 36.5 & 30.8 \\
w/
& 63.2 & 74.7 & 33.5 & 30.0 & 33.2 & 20.3 & 38.0 & 31.1 \\
\hline
\end{tabular}%
}
\caption{
ICA under the O365+GoldG setting on LVIS-minival (steps C$\rightarrow$D, ConvNeXt-B @ 640$\times$640).
}
\label{tab:ica_ablation}
\end{table}

%% file: tables/ablation_efficiency.tex
\begin{table}[!t]
\centering
\small
\renewcommand{\arraystretch}{1.1}
\setlength{\tabcolsep}{4pt}
\resizebox{\columnwidth}{!}{%
\begin{tabular}{clcccc}
\hline
\textbf{ID} & \textbf{Evolution Step} & \textbf{Params} & \textbf{Input} & \textbf{GFLOPs} & \textbf{FPS\textsubscript{gen}} \\
\hline
A & Baseline (Swin-T) & 115M & 800$\times$1333 & 257 & 15.4 \\
B$_1$ & + DINOv3-ConvNeXt-B & 173M & 800$\times$1333 & 439 & 11.6 \\
B$_2$ & + Reduce to 640$\times$640 & 173M & 640$\times$640 & 208 & 19.4 \\
B$_3$ & + Hybrid encoder & 172M & 640$\times$640 & 164 & 22.0 \\
C & + Prompt-aware dec. & 174M & 640$\times$640 & 165 & 21.3 \\
D & + ICA & 174M & 640$\times$640 & 165 & 21.3 \\
E & + Data engine & 174M & 640$\times$640 & 165 & 21.3 \\
\hline
\end{tabular}%
}
\caption{
Component-wise Params, GFLOPs, and $\mathrm{FPS}_{\mathrm{gen}}$.
B$_1$ to B$_3$ factorize step~B in Table~\ref{tab:component_wise_ablation}; C to E match the same table.
Accuracy under each prompting mode is reported in Table~\ref{tab:component_wise_ablation}.
}
\label{tab:component_wise_efficiency}
\end{table}

%% file: tables/ablation_data_scaling.tex
\begin{table*}[!t]
\centering
\small
\setlength{\tabcolsep}{4.5pt}
\renewcommand{\arraystretch}{1.15}
\begin{tabular}{lccccccc}
\hline
\textbf{Model} & \textbf{Training data}
& \multicolumn{3}{c}{\textbf{COCO}}
& \multicolumn{3}{c}{\textbf{LVIS-mv}} \\
\cline{3-8}
& & Visual-I & Visual-G & Text & Visual-I & Visual-G & Text \\
\hline
T-Rex2-T & official
& 56.6 & 38.8 & 45.8 & 59.3 & 37.4 & 42.8 \\
\hline
T-Rex2-T (re-impl.)
& Base
& 57.2 & 41.3 & 45.8 & 56.3 & 33.6 & 26.5 \\
& Mid
& 56.6 & 42.1 & 47.3 & 57.2 & 36.6 & 39.0 \\
& Full
& 56.0 & 41.4 & 47.1 & 59.4 & 40.5 & 45.2 \\
\hline
OPUS (Swin-T)
& Base
& 61.2 & 37.6 & 45.2 & 64.1 & 34.4 & 33.9 \\
& Full
& 68.1 & 35.8 & 47.1 & 70.6 & 35.8 & 44.6 \\
\hline
OPUS (Ours)
& Base
& 64.6 & 43.0 & 48.2 & 63.2 & 33.5 & 33.2 \\
& Full
& 68.1 & 43.4 & 49.6 & 69.2 & 38.6 & 43.0 \\
\hline
\end{tabular}
\caption{
Data scaling on COCO and LVIS-minival.
Base: O365$^{\mathrm{v}}$~\citep{O365}+GoldG~\citep{GoldG}.
Mid: Base+OI$^{\mathrm{v}}$~\citep{OpenImages}+V3Det~\citep{V3det}+HT$^{\mathrm{v}}$~\citep{HierText}+CH$^{\mathrm{v}}$~\citep{CrowdHuman}+Bamboo$^{\mathrm{v}\dagger}$~\citep{Bamboo}.
Full: Mid+CC3M$^\dagger$~\citep{CC3M}+SA-1B$^{\mathrm{v}\dagger}$~\citep{SA-1B} (7.98M images).
$^{\mathrm{v}}$: visual-prompt training (all sources used for text); $^\dagger$: SAM3-based data-engine annotations.
Top: progressive scaling on T-Rex2-T (re-impl.); official T-Rex2-T is shown for reference.
Bottom: cross-framework scaling under the same Base and Full sets.
}
\label{tab:data_scaling}
\end{table*}

%% file: sections/5_conclusion.tex
\section{Conclusion}
\label{sec:conclusion}

We presented OPUS, a simple unified framework for open-vocabulary detection.
OPUS simplifies unified OVD along three axes: a semantic-rich DINOv3 backbone with efficient hybrid encoding and a prompt-aware decoder that avoids prompt-specific branches for unified prompt reasoning, a one-stage text-visual training strategy with Instance-level Contrastive Alignment, and a SAM3-based single-pass data engine.
Across COCO, LVIS-minival, and ODinW35, OPUS achieves state-of-the-art Visual-I performance while maintaining strong Text and Visual-G accuracy, and further turns mixed prompting from interference into genuine complementarity.

Our analysis reveals that this complementarity evolves with grounding supervision.
As Text prompting recovers long-tailed semantic coverage, Visual-G increasingly relies on textual semantics in the mixed setting, exposing the limitation of noisy cross-image category prototypes.
This suggests that improving generic visual prompting through better reference selection, prototype aggregation, and prototype denoising, is an important direction for future unified OVD systems.
Overall, OPUS demonstrates that simplicity and strong unified prompting capability can be achieved together in open-vocabulary detection.


%% file: sections/6_appendix.tex
\section{Method Details}
\label{appendix:method_details}
\input{tables/appendix_A}
\subsection{Model Architecture and Forward}
\label{sec:app-model}

Algorithm~\ref{alg:train-forward} summarizes the forward propagation and loss computation of our unified detector during training. Given an input image, the model first encodes category prompts using the text encoder and extracts multi-scale visual features through the backbone and hybrid encoder. Under the \textsc{visual} mode, the Visual Prompt Encoder (VPE) further produces category-aware visual prompt embeddings and merges them into the text prompt space through visual-to-text prompt interaction. The decoder adopts alternating self-attention, text cross-attention, and image cross-attention for iterative query refinement. Following the two-stage query initialization and denoising strategy, the model computes both matching and denoising detection losses. Additional alignment and instance-level contrastive objectives are introduced only for visual prompt training.

\subsection{Training and Optimization}
\label{sec:app-train}

Algorithm~\ref{alg:train-loop} presents the alternating optimization strategy used in training. To jointly support text prompting and visual prompting learning under a unified detector, we alternate between visual batches and text prompt batches with a ratio of $1{:}8$. Text batches optimize conventional open-vocabulary detection objectives using enhanced category prompts and sampled negative categories, while visual-prompt batches additionally enable visual-text prompt interaction and auxiliary visual prompt losses. All parameters are optimized end-to-end using AdamW.

\subsection{Inference Workflows}
\label{sec:app-infer}

Algorithm~\ref{alg:inference} describes the unified inference workflow supporting four inference modes: text prompting, interactive visual prompting, generic visual prompting, mixed prompting. 
Depending on the inference mode, the detector dynamically constructs textual prompts, user-provided interactive prompts, or cached visual embeddings. The encoded visual prompts are merged into the text embedding space through the same visual-to-text interaction used during training.
For Mixed prompting, there is no learned fusion module: text and class-level visual embeddings are averaged at matched class slots, identical to the training-time merge.
The decoder then performs iterative object query refinement to produce final open-vocabulary detections. For generic visual prompting, visual embeddings can be precomputed and cached to improve inference efficiency.

\subsection{Data Engine}
\label{appendix:data-engine}
Section~\ref{sec:method} describes our SAM3-based single-pass pipeline that converts text-image, classification, and segmentation sources into Bamboo-CLS, CC3M, and annotated SA-1B supervision.
Here we detail the filtering stages and show qualitative cases.
Dataset statistics are summarized in Table~\ref{tab:datasets}.

\paragraph{Human verification.}
\input{tables/ablation_data_engine_quality}
We randomly sample 200 images from each generated source and manually verify every pseudo box--phrase pair after filtering.
A box counts as incorrect if localization or the assigned label is wrong.
Table~\ref{tab:data_engine_quality} reports box accuracies near 90\% on Bamboo-CLS, CC3M, and SA-1B.

\paragraph{Source-specific preprocess filtering.}
Different sources introduce different noise.
For CC3M, noun phrases extracted from captions are used only as candidate text prompts and must be visually grounded by SAM3, which removes many ungroundable phrases.
For Bamboo-CLS, image-level labels can be coarser than object regions or cover only part of the visible content; subsequent grounding and filtering mitigate this mismatch while retaining scalable supervision.
For SA-1B, semantic labels come from TAP over a vocabulary of 2560 common categories; we discard unreliable tags with category-dependent confidence thresholds.

\paragraph{Unified post-processing.}
All generated boxes then pass through a shared post-processing stage: confidence thresholding, category-wise and image-wise NMS, removal of near full-image boxes, and filtering of large boxes that enclose many small same-category instances.

\begin{figure*}[t]
\centering
\includegraphics[width=\textwidth]{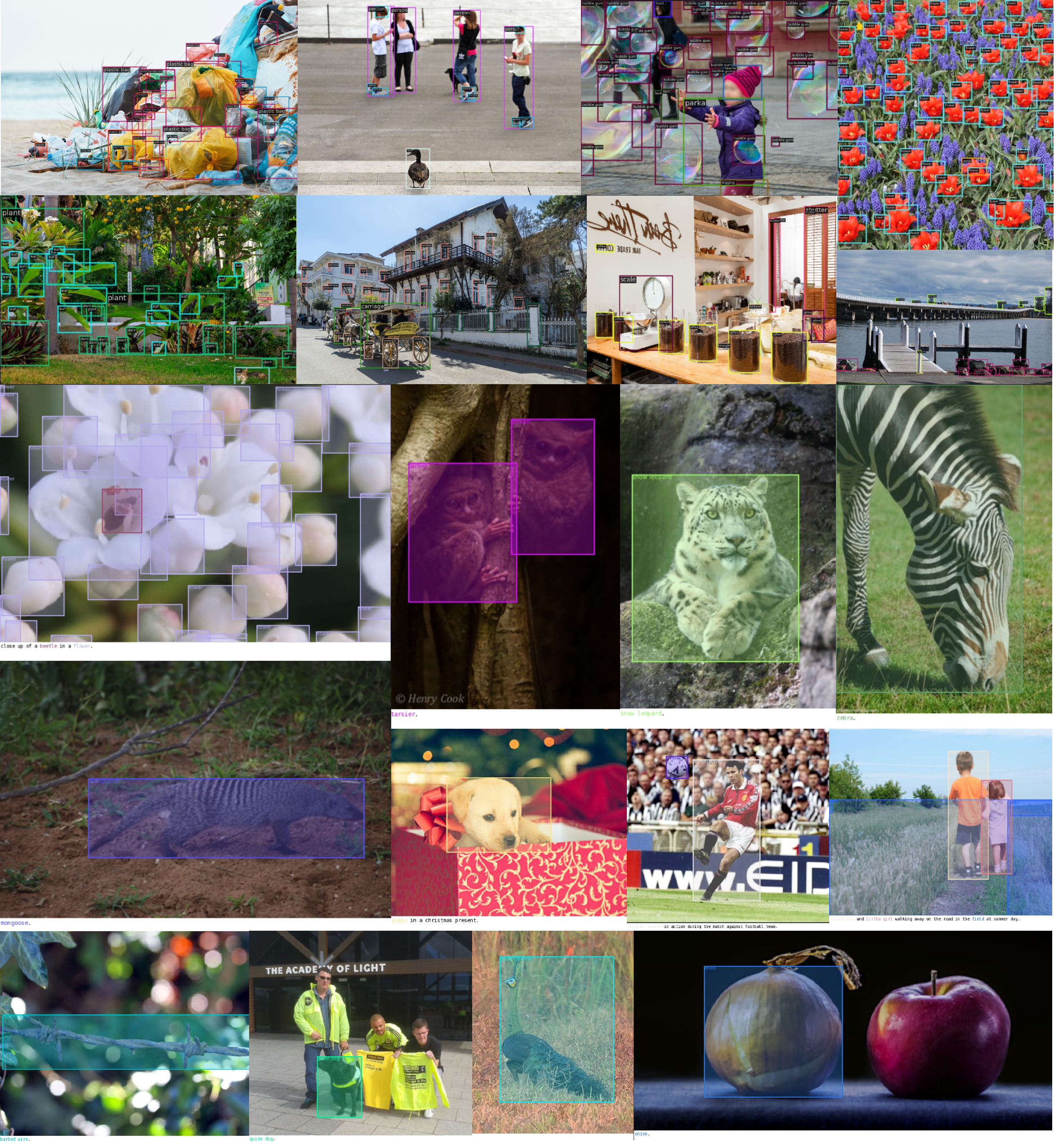}
\caption{Representative SAM3 pseudo-label success cases after filtering on Bamboo-CLS, CC3M, and SA-1B.}
\label{fig:data-engine-success}
\end{figure*}

\begin{figure*}[t]
\centering
\includegraphics[width=\textwidth]{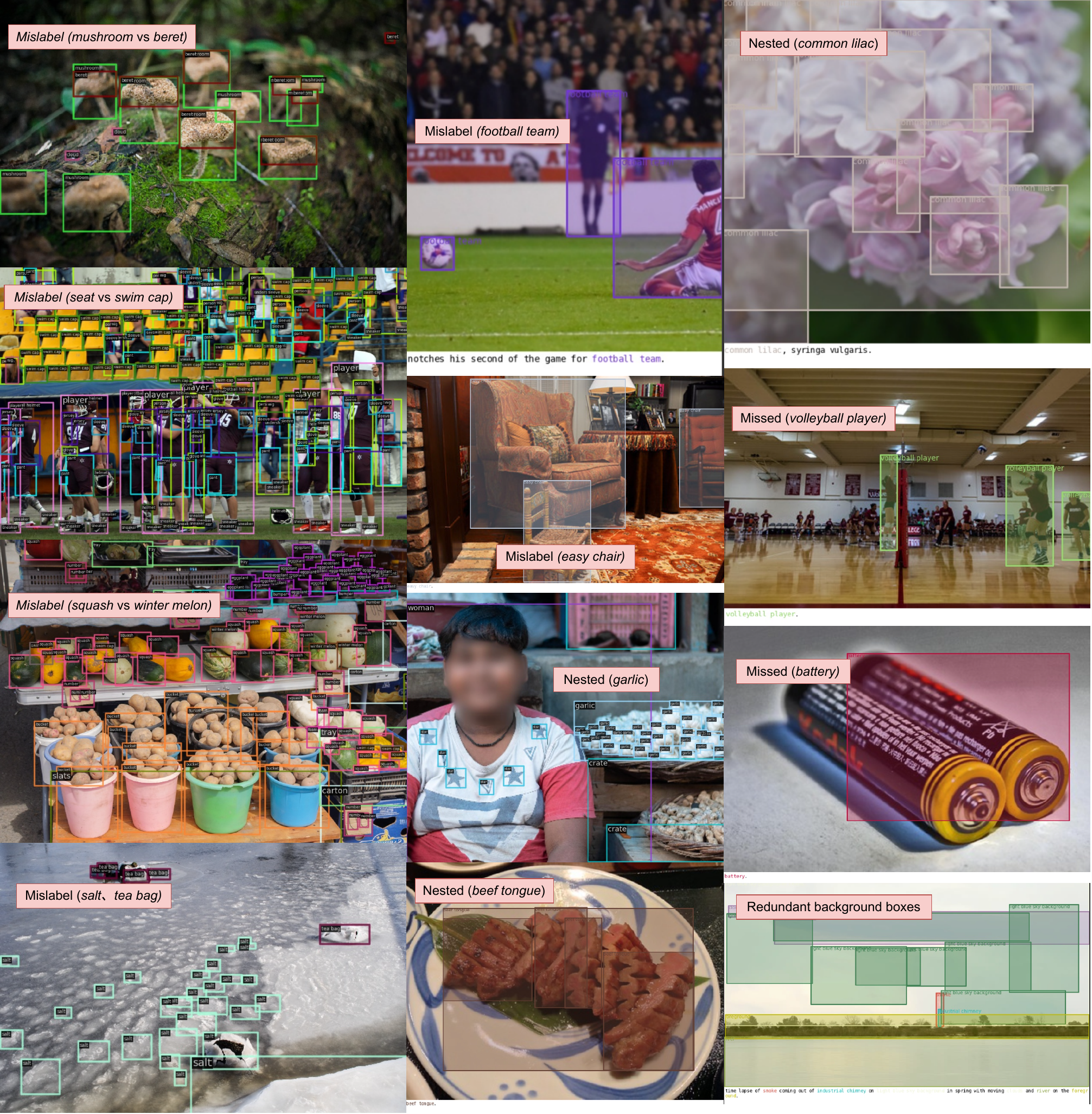}
\caption{Representative SAM3 pseudo-label failure cases after filtering on Bamboo-CLS, CC3M, and SA-1B (mislabel, nested, and redundant background boxes).}
\label{fig:data-engine-failure}
\end{figure*}

\paragraph{Qualitative cases.}
Figure~\ref{fig:data-engine-success} and Figure~\ref{fig:data-engine-failure} show representative success and failure cases after filtering, with one column each for Bamboo-CLS, CC3M, and SA-1B.
Successful examples typically produce tight object-level boxes with consistent labels.
Residual failures include mislabeled boxes, nested boxes, and redundant background boxes on regions without clear object features (e.g., sky).
On Bamboo-CLS and CC3M, nested or redundant boxes are more common; on SA-1B, mislabels appear more often.

\section{Experimental Setup Details}
\label{appendix:experimental_details}

\subsection{Training Hyperparameter Configuration}
\label{appendix:training_details}
\input{tables/appendix_train_config}
Our training hyperparameter configuration is summarized in Table~\ref{tab:train-config}. Unless otherwise specified, all models are trained under the same optimization setting, including input resolution, learning-rate schedule, alternating training ratio, and loss weighting strategy, to ensure fair comparison under a fixed efficiency budget.

\subsection{Sensitivity Analysis}
\label{appendix:sensitivity}

We study sensitivity to Visual-G reference sampling ($N$), the visual:text alternating ratio, and ICA temperature $\tau$.

\subsubsection{Visual-G Reference Sampling}

\input{tables/ablation_visual_g_ref}
Table~\ref{tab:ref-n} varies $N \in \{1,4,8,16,32,64\}$ to verify the default $N{=}16$ choice.
Visual-G needs enough references for stable category-level prototypes, rising from 25.0 AP at $N{=}1$ to 38.6 AP at $N{=}16$ on LVIS-minival, and then saturates beyond $N{=}16$.
For rare categories, the training set may contain fewer usable images than $N$, so larger $N$ does not always mean more distinct references.
Under Mixed prompting, text provides an additional anchor, so performance varies less with $N$ and already exceeds Text-only from $N{=}4$ on LVIS-minival and from $N{=}16$ on COCO.
Five random seeds at the default $N{=}16$ confirm that this choice is stable: Visual-G is $43.3{\pm}0.1$ on COCO and $38.6{\pm}0.3$ on LVIS-minival, and Mixed is $49.9{\pm}0.1$ and $45.5{\pm}0.2$.

\subsubsection{Visual:Text Alternating Ratio}

\input{tables/appendix_batch_ratio}
Table~\ref{tab:batch-ratio} ablates the visual:text alternating ratio on the ICA-equipped model under the O365+GoldG setting (main-text Table~\ref{tab:component_wise_ablation}, step~D).
A larger visual share (1:4) favors Visual-I, while a larger text share (1:16) favors Text, Visual-G, and Mixed.
We adopt 1:8 as a balanced default across all four prompt modes, matching the main experiments.

\subsubsection{ICA Temperature}

\input{tables/appendix_ica_temp}
Table~\ref{tab:ica-temp} varies the ICA temperature $\tau$ under the same setting as Table~\ref{tab:batch-ratio}, with $w_{\mathrm{ICA}}{=}1.0$.
Across $\tau \in \{0.05,0.07,0.10\}$, Visual-I, Text, and Mixed change only slightly, and Visual-G stays within $0.3$ AP.
We therefore keep $\tau{=}0.07$ as the default.

\subsection{Evaluation Protocol}
\label{appendix:eval_details}
\subsubsection{Accuracy Evaluation Protocols}

\paragraph{Text Protocol.}
Under the text prompt protocol, category names are used as textual inputs and encoded by a pretrained text encoder.
The resulting text embeddings are projected into the shared prompt space and used as category prompts for open-vocabulary detection.
This protocol evaluates the detector's ability to recognize categories from language descriptions alone.

\paragraph{Visual-I Protocol.}
Under the interactive visual prompt protocol, one ground-truth instance is randomly selected for each category in a test image and used as the visual prompt.
The model then detects all instances of the corresponding categories in the same image.
This setting simulates interactive applications such as annotation and object counting, where users provide instance-level guidance at inference time.

\paragraph{Visual-G Protocol.}
Under the generic visual prompt protocol, category-level visual prompts are constructed offline from reference images.
For each category, we randomly sample a fixed number of training images, encode their visual prompts, and aggregate them into a category-level visual prototype.
The resulting prototypes are reused across all test images during inference.
This protocol evaluates whether visual prompts can serve as generic category descriptions without per-image interaction.

\paragraph{Mixed Protocol (Text + Visual-G).}
Under the mixed protocol, text prompt embeddings and Visual-G prototypes are independently constructed and jointly used as category prompts during inference.
The two embeddings are averaged at matched class slots without a learned fusion module.
This setting evaluates whether semantic text cues and appearance-level visual cues can complement each other in a unified detector.

\subsubsection{Efficiency Evaluation Protocols}

\paragraph{Interactive Speed ($\mathrm{FPS}_{\mathrm{inter}}$).}
This metric measures inference throughput in the interactive visual prompting setting.
Following prior work, image features are cached for each test image, and the measured runtime includes visual prompt encoding and decoder inference for each user interaction.
This reflects the latency-sensitive use case where users iteratively provide visual prompts on the same image.

\paragraph{Generic Speed ($\mathrm{FPS}_{\mathrm{gen}}$).}
This metric measures inference throughput for Text and Visual-G prompting.
In this setting, category prompt embeddings are precomputed offline and reused across test images.
The measured runtime therefore includes image feature extraction and decoder inference, but excludes repeated prompt encoding.
This reflects standard generic detection, where a fixed set of category prompts is applied to many images.

\section{More Experimental Analysis}
\label{appendix:more_experimental_analysis}

\subsection{Breakdown of Component-wise Ablation}
\label{appendix:breakdown_component_analysis}

\input{tables/appendix_detailed_ablation_lvis}
\input{tables/appendix_efficiency_full}

Table~\ref{tab:detailed_ablation_lvis} keeps the same A$\to$E steps as Table~\ref{tab:component_wise_ablation}, and breaks each step into AP$_f$/AP$_c$/AP$_r$.
Overall, the progression from the baseline to the full OPUS model improves performance across most category groups and prompting modes, showing that the gains are not limited to frequent categories.

A notable exception appears when adding the prompt-aware decoder in Row~C.
While the decoder improves Text and Visual-I prompting, it temporarily reduces Visual-G performance, with the largest drop on rare categories.
This supports our interpretation that Visual-G is more sensitive to the mismatch between Visual-I-style training prompts and cross-image category prototypes used during Visual-G inference.
Rare categories are particularly affected because their prototypes are aggregated from fewer and more diverse examples, making them less stable than frequent-category prototypes.

Adding ICA in Row~D produces a selective recovery pattern under Visual-G.
Although the overall Visual-G AP recovers only modestly, AP$_r$ improves substantially from 26.1 to 30.0, while AP$_f$ gains slightly and AP$_c$ remains below the pre-decoder level.
This pattern suggests that ICA is particularly useful when category-level visual prototypes are unreliable, especially for rare categories.

To understand this effect, recall that the base alignment objective $\mathcal{L}_{\mathrm{align}}$ operates at the class level.
It trains the aggregated visual prompt feature $\mathbf{P}^{v}_{\mathrm{cls}}$ by pulling it toward the corresponding text prompt feature $\mathbf{P}^{t}$.
While this alignment encourages text-visual category consistency, it also makes Visual-G prototypes depend on a global semantic anchor.
Such an anchor is coarse by design: text features mainly encode category-level semantics rather than instance-level appearance variations.
This limitation becomes more pronounced for rare categories, where both the text representations and the cross-image visual prototypes receive less reliable supervision.
As a result, aligning Visual-G features primarily through $\mathcal{L}_{\mathrm{align}}$ may weaken fine-grained visual details and make rare-category prototypes inherit the limitations of the text modality.

ICA addresses this issue from a different direction.
Instead of aligning only the aggregated class-level visual prototype to text, ICA aligns instance-level visual prompt features $\mathbf{P}^{v}_{\mathrm{con}}$ directly with their matched decoder queries.
These decoder queries are grounded in image regions through detection supervision, and therefore provide a more localized and appearance-sensitive target than text embeddings.
This instance-query alignment helps preserve fine-grained visual information without being constrained solely by the semantic coarseness of the text anchor.
It also provides a stronger learning signal for rare categories, where class-level text-visual alignment is least reliable.
This explains why ICA yields its clearest Visual-G recovery on AP$_r$ rather than uniformly improving all frequency groups.

\subsection{Component-wise Efficiency and Accuracy}
\label{appendix:efficiency}

Table~\ref{tab:component_wise_efficiency_full} lists Params, input size, GFLOPs, and FPS together with Visual-I/Visual-G/Text/Mixed AP for each intermediate configuration.
It complements the main-text efficiency table, which omits AP to avoid duplicating Table~\ref{tab:component_wise_ablation}, and makes the accuracy/efficiency trade-off along step~B easier to inspect in one place.

\subsection{Comparison with Staged Visual-Prompt Training}
\label{appendix:staged_trex2}
\input{tables/appendix_staged_trex2}
Table~\ref{tab:data_scaling_trex2} recalls official T-Rex2-T second-stage visual-prompt results from T-Rex2 Table~6.
These numbers are a reference comparison against reported results, not a controlled re-training under the same implementation.

With only O365 for visual-prompt training, our one-stage re-impl.\ already reaches Visual-I 57.2/56.3 on COCO/LVIS-minival (Table~\ref{tab:data_scaling}), while official staged training reports 41.1/40.6 before SA-1B and 56.6/59.3 after.
Under Visual-G, adding SA-1B lowers official T-Rex2-T from 41.1 to 38.8 on COCO and from 38.1 to 37.4 on LVIS-minival.
The one-stage re-impl.\ instead stays flat on COCO (41.3$\rightarrow$41.4) and rises on LVIS-minival (33.6$\rightarrow$40.5).

\subsection{Visualizations}
\label{appendix:visualizations}

We provide qualitative results for interactive visual prompting, generic visual prompting, and text prompting.
These examples complement the quantitative results by showing how OPUS behaves across dense scenes, cross-image exemplar transfer, and open-vocabulary text-guided detection.

\begin{figure*}[t]
  \centering
  \includegraphics[width=\textwidth]{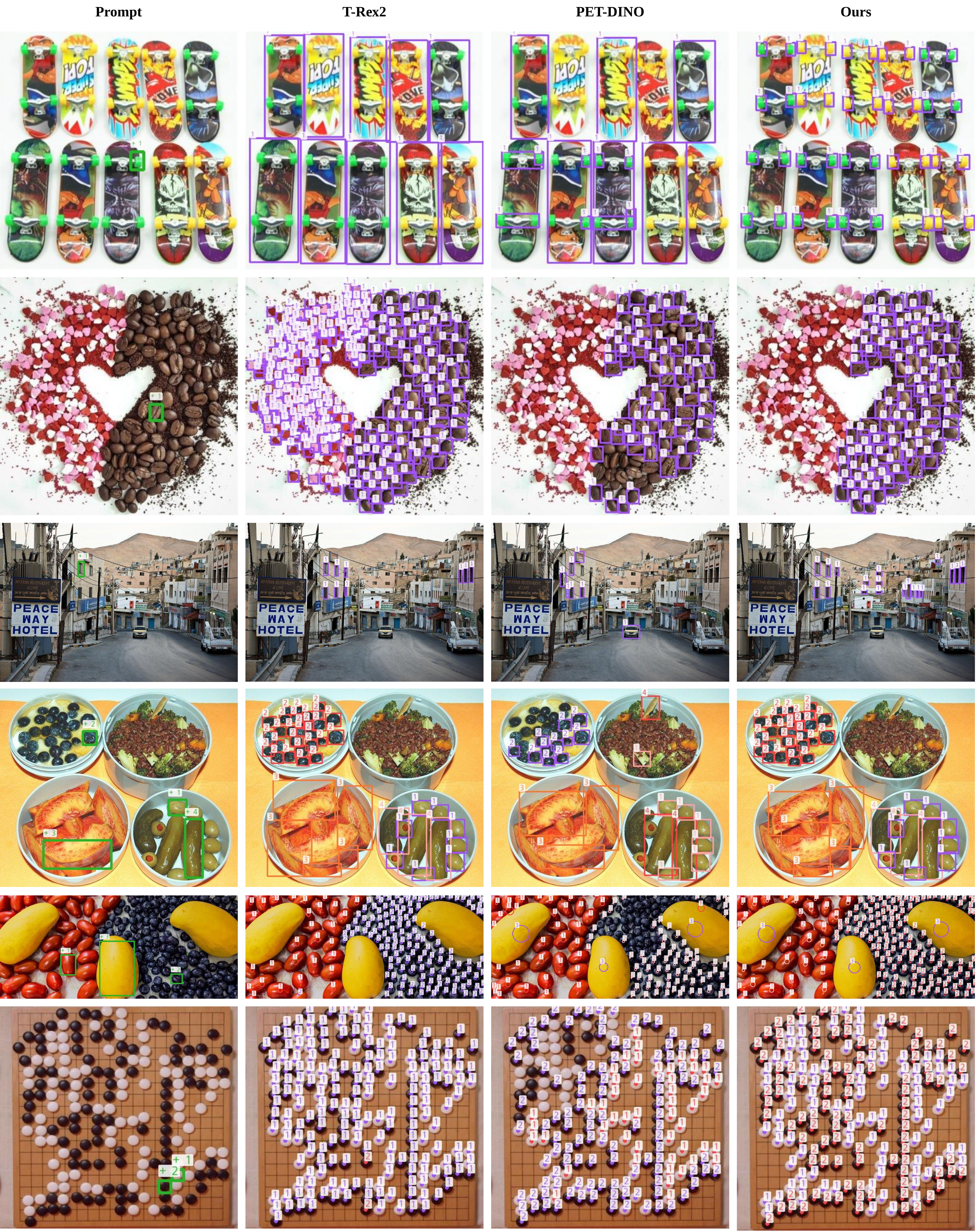}
  \caption{
   Interactive visual prompt detection results of \textbf{OPUS} compared with T-Rex2 and PET-DINO. OPUS demonstrates stronger visual prompt understanding for both single-category and multi-category targets across diverse and dense scenes.
  }
  \label{fig:interactive_dense}
\end{figure*}

Figure~\ref{fig:interactive_dense} compares OPUS with T-Rex2 and PET-DINO under interactive visual prompting.
OPUS demonstrates stronger visual prompt understanding for both single-category and multi-category targets in diverse and dense scenes.

\begin{figure*}[t]
  \centering
  \includegraphics[width=\textwidth]{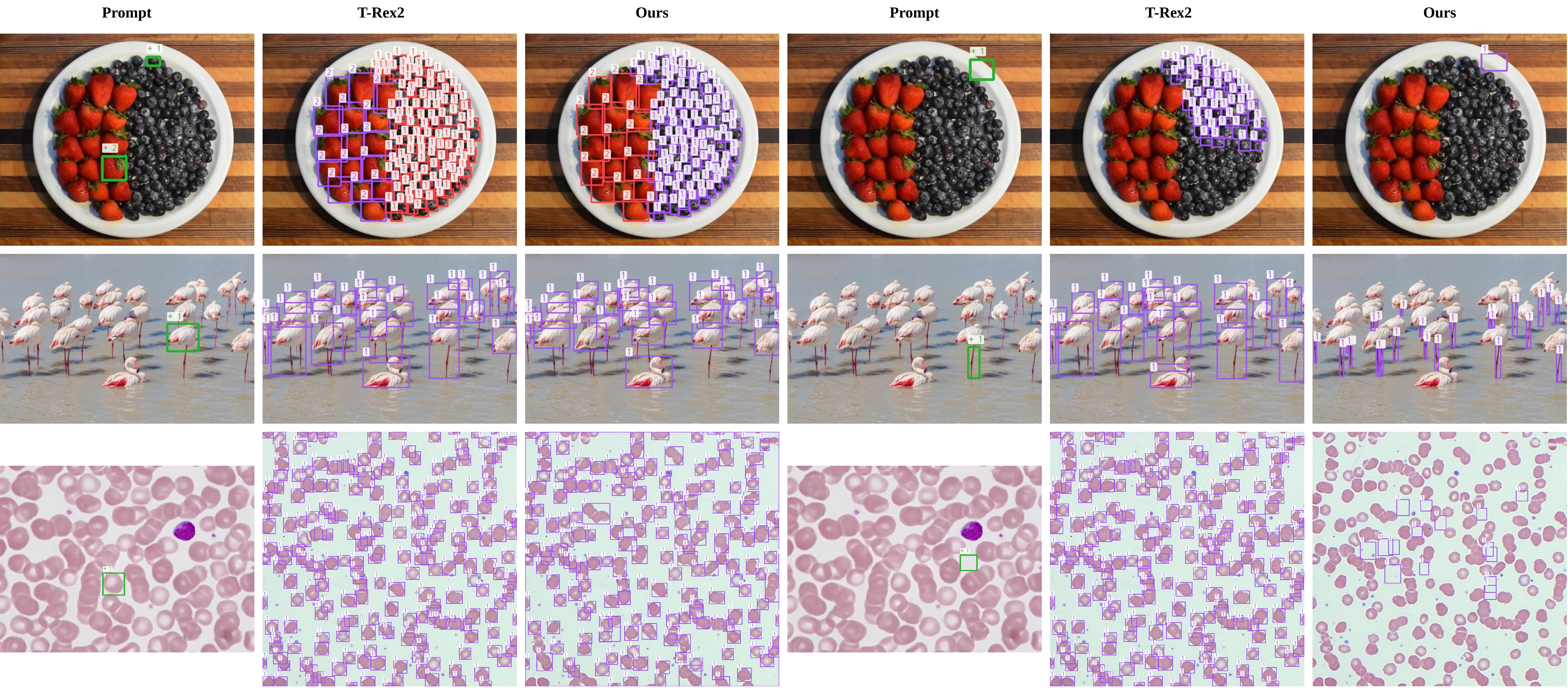}
  \caption{
  Comparison with T-Rex2 on challenging interactive visual detection cases. \textbf{OPUS} produces more accurate localization results with fewer FPs.
  }
  \label{fig:interactive_comparison}
\end{figure*}

Figure~\ref{fig:interactive_comparison} further compares OPUS with T-Rex2 on challenging interactive visual prompting cases.
OPUS produces fewer background false positives and more accurate instance localization.

\begin{figure*}[t]
  \centering
  \includegraphics[width=\textwidth]{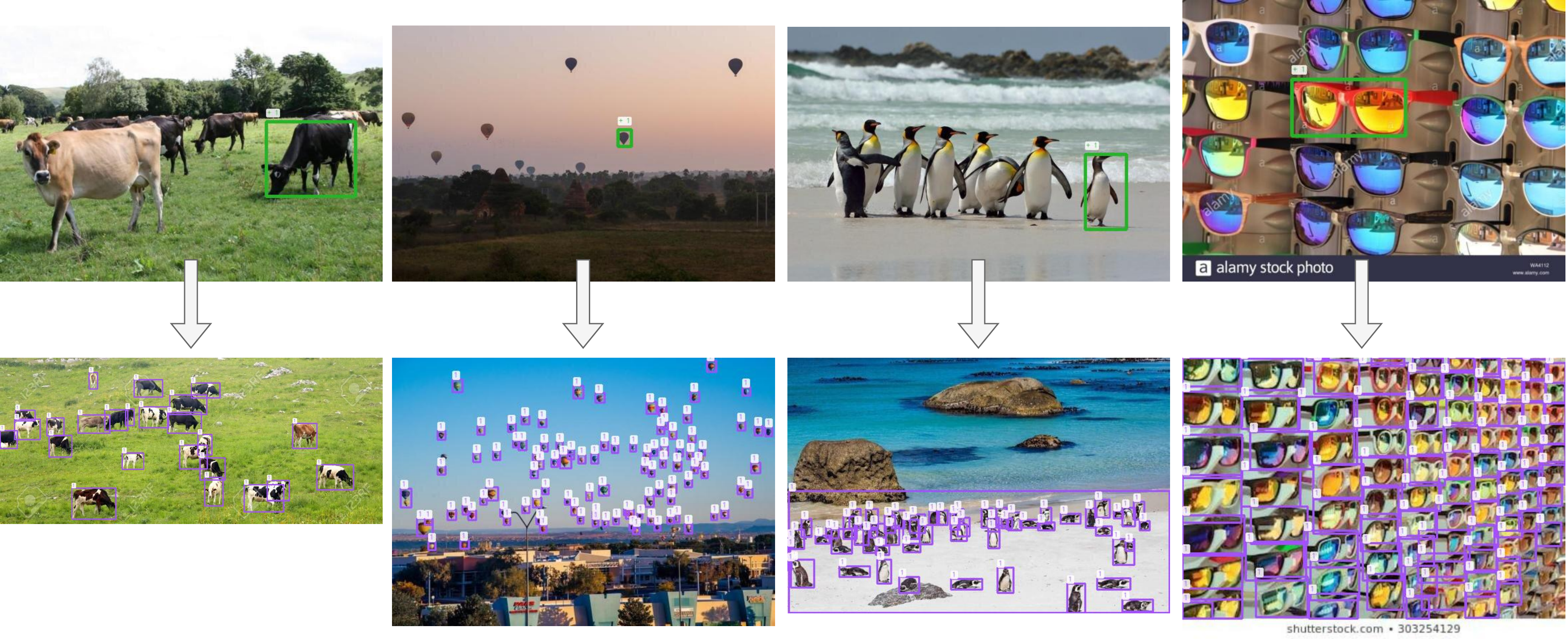}
  \caption{
   Zero-shot detection visualizations of \textbf{OPUS} using cross-image
exemplar visual prompts. The top row shows exemplar prompts,
and the bottom row presents the corresponding detection results.
  }
  \label{fig:generic_visual}
\end{figure*}

Figure~\ref{fig:generic_visual} shows generic visual prompting results, where exemplar prompts from reference images are transferred to different test images.
OPUS effectively preserves category-level visual semantics across changes in scene layout, object scale, and instance appearance.

\begin{figure*}[t]
  \centering
  \includegraphics[width=\textwidth]{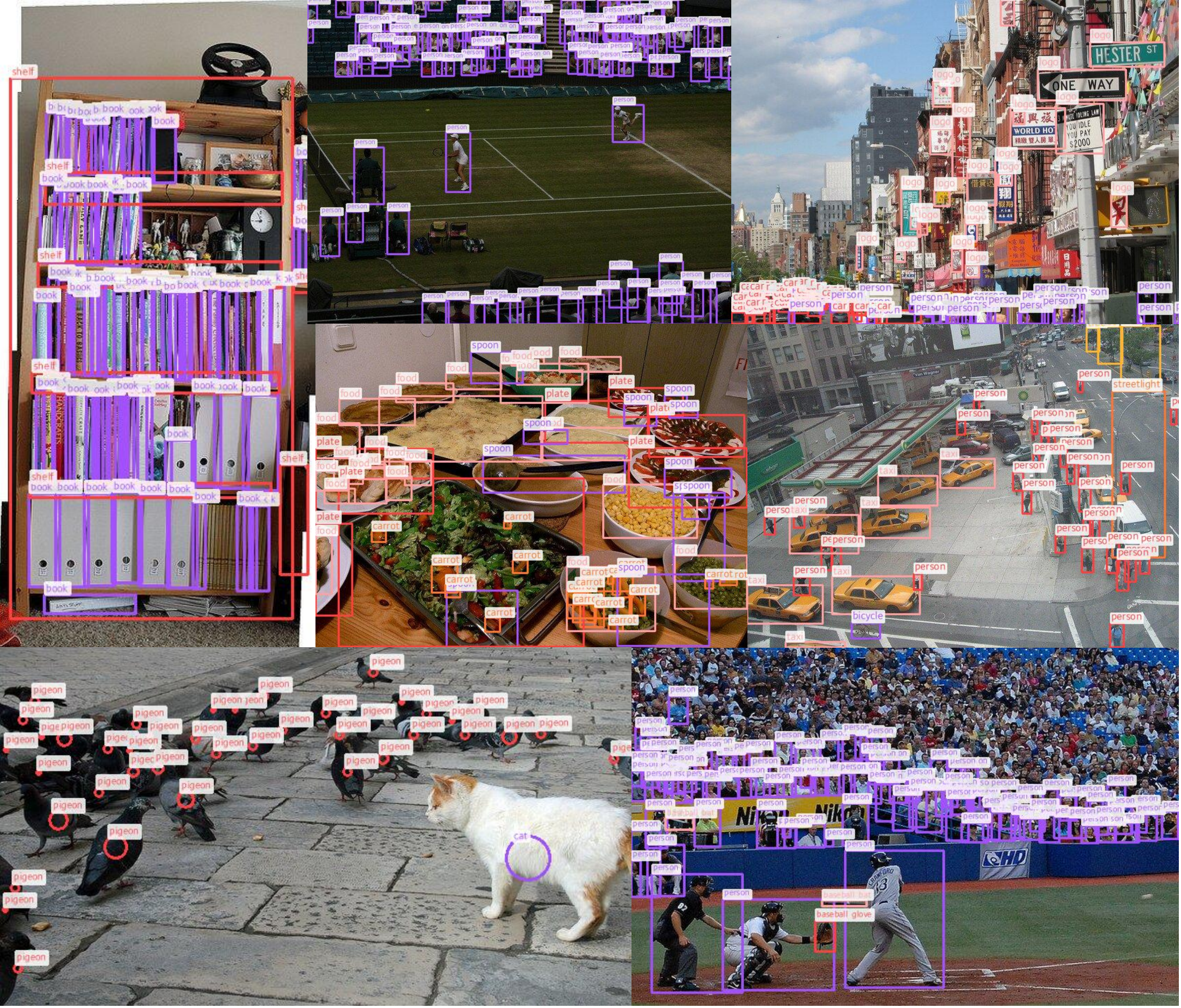}
  \caption{
  Zero-shot detection visualizations of \textbf{OPUS} using text prompts.
  }
  \label{fig:generic_text}
\end{figure*}

Figure~\ref{fig:generic_text} presents text-prompted zero-shot detection examples.

Together, these visualizations show that OPUS provides robust prompt understanding across text, interactive visual, and generic visual prompting settings.

%% file: tables/appendix_A.tex
\begin{algorithm}[t]
\caption{Unified Forward and Training Objective}
\label{alg:train-forward}
\small
\begin{algorithmic}[1]

\REQUIRE Image $I$, annotations $\mathcal{Y}$,
  text prompts $P^{t}$, visual prompts $V$,
  mode $m \in \{\textsc{text}, \textsc{visual}\}$
\ENSURE $\mathcal{L}=\mathcal{L}_{\text{det}}+\mathcal{L}_{\text{align}}+\mathcal{L}_{\text{ica}}$

\STATE $\{F_\ell\}_{\ell \in \{3,4,5\}} \leftarrow \textsc{Backbone}(I)$
\STATE $\{F_\ell\} \leftarrow \textsc{ChannelMapper}(\{F_\ell\})$
\STATE $F \leftarrow \textsc{HybridEncoder}(\{F_\ell\})$

\IF{$m=\textsc{visual}$}
    \STATE $(P_{\text{cls}}^{v}, P_{\text{con}}^{v})
      \leftarrow \textsc{VPE}(F, V)$
    \STATE $(P,\mathcal{A}) \leftarrow
      \textsc{MergeVisualToText}(P^{t}, P_{\text{cls}}^{v})$
\ELSE
    \STATE $P \leftarrow P^{t}$
\ENDIF

\STATE $(Q_0,R_0) \leftarrow \textsc{TopKProposals}(F,P,N_q)$
\STATE $Q_0 \leftarrow \textsc{Concat}(\textsc{BuildDNQueries}(\mathcal{Y},P), Q_0)$

\FOR{$l=1$ to $6$}
    \STATE $Q_l \leftarrow \textsc{SelfAttn}(Q_{l-1})$
    \STATE $Q_l \leftarrow \textsc{CrossAttnPrompt}(Q_l,P)$
    \STATE $Q_l \leftarrow \textsc{CrossAttnImage}(Q_l,F,R_{l-1})$
    \STATE $Q_l \leftarrow \textsc{FFN}(Q_l)$
    \STATE $R_l \leftarrow \textsc{RefineReferencePoints}(Q_l,R_{l-1})$
\ENDFOR

\STATE Split $Q_l$ into $Q^{\text{match}}, Q^{\text{dn}}$
\STATE $(\hat{y},\hat{b}),(\hat{y}_{\text{dn}},\hat{b}_{\text{dn}})
  \leftarrow \textsc{DetHead}(Q^{\text{match}}, Q^{\text{dn}}, P)$
\STATE $\Pi \leftarrow \textsc{HungarianMatch}(\hat{y},\hat{b},\mathcal{Y})$
\STATE $\mathcal{L}_{\text{det}}\leftarrow
  \textsc{DetLoss}(\hat{y},\hat{b},\hat{y}_{\text{dn}},\hat{b}_{\text{dn}},\mathcal{Y},\Pi)$

\IF{$m=\textsc{visual}$}
    \STATE $\mathcal{L}_{\text{align}} \leftarrow
      \textsc{AlignLoss}(P^{t}, P_{\text{cls}}^{v};\mathcal{A})$
    \STATE $\mathcal{L}_{\text{ica}} \leftarrow
      \textsc{InfoNCE}(Q^{\text{match}}, P_{\text{con}}^{v}, \Pi)$
\ELSE
    \STATE $\mathcal{L}_{\text{align}}, \mathcal{L}_{\text{ica}} \leftarrow 0$
\ENDIF

\RETURN $\mathcal{L}_{\text{det}}
  +\mathcal{L}_{\text{align}}+\mathcal{L}_{\text{ica}}$
\end{algorithmic}
\end{algorithm}

\begin{algorithm}[t]
\small
\caption{Alternating Training}
\label{alg:train-loop}

\begin{algorithmic}[1]
\REQUIRE text dataloader $\mathcal{D}_{\text{text}}$,
visual dataloader $\mathcal{D}_{\text{vis}}$,
total iterations $T_{\max}$,
alternating ratio $1{:}8$
(\textsc{visual} : \textsc{text})
\ENSURE Parameters $\Theta$

\FOR{$t = 0$ \TO $T_{\max}-1$}

    \IF{$t \bmod 9 = 8$}
        \STATE $(I, \mathcal{Y}) \sim \mathcal{D}_{\text{vis}}$
        \STATE $\textit{mode} \leftarrow \textsc{visual}$
    \ELSE
        \STATE $(I, \mathcal{Y}) \sim \mathcal{D}_{\text{text}}$
        \STATE $\textit{mode} \leftarrow \textsc{text}$
    \ENDIF

    \STATE $\mathcal{L} \leftarrow
    \textsc{ForwardAndLoss}(I,\mathcal{Y},\textit{mode};\Theta)$
    \COMMENT{Algorithm~\ref{alg:train-forward}}

    \STATE $\Theta \leftarrow
    \textsc{AdamWUpdate}(\Theta,\nabla_{\Theta}\mathcal{L})$

\ENDFOR

\end{algorithmic}
\end{algorithm}

\begin{algorithm}[t]
\small
\caption{Open-Vocabulary Inference}
\label{alg:inference}
\begin{algorithmic}[1]

\REQUIRE input image~$I$; mode~$m$; text prompts~$P^{t}$; visual prompts~$V$
\ENSURE detections $\{(b_j, s_j, c_j)\}_{j=1}^{N}$

\STATE $F \leftarrow \textsc{Encoder}(I)$

\IF{$m = \textsc{text}$}
    \STATE $P \leftarrow P^{t}$
\ELSE
    \STATE $(P_{\text{cls}}^{v}, P_{\text{con}}^{v}) \leftarrow
      \textsc{VPE}(F, V)$
      \textbf{ or } precomputed $P_{\text{cls}}^{v}$
    \STATE $P \leftarrow \textsc{MergeVisualToText}^{\mathrm{infer}}
      (P^{t}, P_{\text{cls}}^{v}, m)$
\ENDIF

\STATE $(Q_0,R_0) \leftarrow \textsc{TopKProposals}(F,P,N_q)$
\STATE $Q,R \leftarrow \textsc{Decoder}(Q_0,P,F,R_0)$
\STATE $\{(b_j,s_j,c_j)\} \leftarrow \textsc{DetHead}.\textsc{Predict}(Q,P)$

\RETURN $\{(b_j,s_j,c_j)\}$

\end{algorithmic}
\end{algorithm}

\begin{table}[!t]
\centering
\resizebox{\columnwidth}{!}{%
\begin{tabular}{lccc}
\hline
\textbf{Dataset} & \textbf{Prompt Type} & \textbf{Images} & \textbf{Ratio} \\
\hline
Objects365       & Text / Visual & 0.6M  & 7.5\%  \\
OpenImages       & Text / Visual & 1.67M & 20.9\% \\
GoldG            & Text          & 76K   & 1.0\%  \\
V3Det            & Text          & 0.15M & 1.9\%  \\
CrowdHuman       & Visual        & 15K   & 0.2\%  \\
HierText         & Visual        & 8K    & 0.1\%  \\
Bamboo-Det       & Text / Visual & 1.08M & 13.5\% \\
\hline
Bamboo-CLS$^\dagger$   & Text   & 1.56M & 19.6\% \\
CC3M$^\dagger$         & Text   & 1.32M & 16.5\% \\
SA-1B$^\dagger$        & Visual & 1.5M  & 18.8\% \\
\hline
Total            & Text / Visual & 7.98M & 100.0\% \\
\hline
\end{tabular}%
}
\caption{Training datasets used in OPUS. Datasets marked with $\dagger$ have annotations generated by our SAM3-based data engine pipeline.}
\label{tab:datasets}
\end{table}

%% file: tables/ablation_data_engine_quality.tex
\begin{table}[!t]
\centering
\small
\setlength{\tabcolsep}{6pt}
\renewcommand{\arraystretch}{1.1}
\begin{tabular}{lccc}
\hline
\textbf{Source} & \textbf{\#Images} & \textbf{\#Boxes} & \textbf{Box Acc.\ (\%)} \\
\hline
Bamboo-CLS & 200 & 653 & 91.9 \\
CC3M & 200 & 1291 & 89.5 \\
SA-1B & 200 & 7897 & 91.2 \\
\hline
\end{tabular}
\caption{
Human verification of SAM3 pseudo-labels after filtering.
A box is counted as incorrect if localization or the assigned label is wrong.
}
\label{tab:data_engine_quality}
\end{table}

%% file: tables/appendix_train_config.tex
\begin{table*}[t]
\centering
\begin{tabular}{ll}
\toprule
\textbf{Parameter} & \textbf{Value} \\
\midrule
Backbone & DINOv3-ConvNeXt-B \\
Text encoder & CLIP ViT-B/32 \\
Image size & $640 \times 640$ \\
Batch size per GPU & 8 \\
GPUs & 16 \\
Global batch size & 128 \\
Training iterations & 400K \\
Alternating ratio &
1:8 (\textsc{visual}:\textsc{text}) \\
Optimizer & AdamW \\
Base learning rate & $4 \times 10^{-4}$ \\
Backbone / text encoder LR &
$4 \times 10^{-5}$ \\
Weight decay & $1 \times 10^{-4}$ \\
Gradient clipping & 0.1 \\
Warmup &
1K iterations, start factor 0.1 \\
LR scheduler &
MultiStep (50\%, 90\%; $\gamma=0.1$) \\
Precision & FP16 (AMP) \\
Detection queries & 900 \\
DN queries & 100 \\
Max categories per image & 256 \\
Loss weights &
cls / bbox / iou / align / ica \\
& 1.0 / 5.0 / 2.0 / 1.0 / 1.0 \\
Align loss &
Cross-entropy w/ learnable log-scale \\
ICA temperature &
InfoNCE ($\tau=0.07$) \\
Validation interval & 20K iterations \\
\bottomrule
\end{tabular}
\caption{Training configuration.}
\label{tab:train-config}
\end{table*}

%% file: tables/ablation_visual_g_ref.tex
\begin{table}[!t]
\centering
\footnotesize
\setlength{\tabcolsep}{3pt}
\renewcommand{\arraystretch}{1.05}
\begin{tabular}{clccccc}
\hline
\textbf{$N$} & \textbf{Prompt} & \textbf{COCO}
& \multicolumn{4}{c}{\textbf{LVIS-mv}} \\
\cline{4-7}
& & AP
& AP & AP$_f$ & AP$_c$ & AP$_r$ \\
\hline
1  & Visual-G & 27.6 & 25.0 & 25.6 & 25.4 & 20.0 \\
4  & Visual-G & 38.6 & 34.5 & 35.1 & 34.8 & 28.8 \\
8  & Visual-G & 41.6 & 37.1 & 38.0 & 36.9 & 33.6 \\
16 & Visual-G & \textbf{43.4} & \textbf{38.6} & \textbf{39.5} & \textbf{38.8} & \textbf{32.8} \\
32 & Visual-G & 44.3 & 39.7 & 40.1 & 40.7 & 32.3 \\
64 & Visual-G & 44.7 & 39.7 & 40.4 & 40.3 & 33.0 \\
\hline
1  & Mixed & 46.2 & 40.5 & 42.6 & 39.4 & 34.2 \\
4  & Mixed & 49.1 & 44.7 & 46.1 & 44.5 & 38.6 \\
8  & Mixed & 49.4 & 45.1 & 46.7 & 44.2 & 39.9 \\
16 & Mixed & \textbf{49.9} & \textbf{45.2} & \textbf{47.1} & \textbf{44.4} & \textbf{38.7} \\
32 & Mixed & 50.2 & 45.4 & 47.2 & 44.8 & 37.8 \\
64 & Mixed & 50.3 & 45.6 & 47.4 & 45.0 & 38.6 \\
\hline
\end{tabular}
\caption{
Visual-G and Mixed prompting vs.\ number of references ($N$) on the final OPUS model.
Text-only: 49.6 AP (COCO) / 43.0 AP (LVIS-minival).
Bold marks the default $N{=}16$.
}
\label{tab:ref-n}
\end{table}

%% file: tables/appendix_batch_ratio.tex
\begin{table}[t]
\centering
\small
\setlength{\tabcolsep}{6pt}
\renewcommand{\arraystretch}{1.1}
\begin{tabular}{lcccc}
\hline
visual:text & Visual-I & Visual-G & Text & Mixed \\
\hline
1:4  & \textbf{64.6} & 31.5 & 31.6 & 36.8 \\
1:8  & 63.2 & 33.5 & 33.2 & 38.0 \\
1:16 & 58.8 & \textbf{34.6} & \textbf{33.7} & \textbf{38.5} \\
\hline
\end{tabular}
\caption{Sensitivity to the visual:text alternating ratio on LVIS-minival.}
\label{tab:batch-ratio}
\end{table}

%% file: tables/appendix_ica_temp.tex
\begin{table}[t]
\centering
\small
\setlength{\tabcolsep}{6pt}
\renewcommand{\arraystretch}{1.1}
\begin{tabular}{lcccc}
\hline
$\tau$ & Visual-I & Visual-G & Text & Mixed \\
\hline
0.05 & 63.3 & 33.7 & 32.7 & 38.4 \\
0.07 & 63.2 & 33.5 & 33.2 & 38.0 \\
0.10 & 63.1 & 33.4 & 33.2 & 38.5 \\
\hline
\end{tabular}
\caption{ICA temperature sensitivity ($\tau$) on LVIS-minival.}
\label{tab:ica-temp}
\end{table}

%% file: tables/appendix_detailed_ablation_lvis.tex
\begin{table*}[t]
\centering
\small
\renewcommand{\arraystretch}{1.4}
\setlength{\tabcolsep}{4pt}
\definecolor{vi}{HTML}{F2F4FA}
\definecolor{vg}{HTML}{EEF7EE}
\definecolor{txt}{HTML}{F7F7F7}
\definecolor{mix}{HTML}{FFF6E8}
\definecolor{hl}{HTML}{FFF176}
\resizebox{\textwidth}{!}{
\begin{tabular}{c|cccc|cccc|cccc|cccc}
\hline
\textbf{ID}
& \multicolumn{4}{c|}{\cellcolor{vi}\textbf{Visual-I}}
& \multicolumn{4}{c|}{\cellcolor{vg}\textbf{Visual-G}}
& \multicolumn{4}{c|}{\cellcolor{txt}\textbf{Text}}
& \multicolumn{4}{c}{\cellcolor{mix}\textbf{Mixed}} \\
\cline{2-17}
& \cellcolor{vi}\textbf{AP} & \cellcolor{vi}\textbf{AP$_f$} & \cellcolor{vi}\textbf{AP$_c$} & \cellcolor{vi}\textbf{AP$_r$}
& \cellcolor{vg}\textbf{AP} & \cellcolor{vg}\textbf{AP$_f$} & \cellcolor{vg}\textbf{AP$_c$} & \cellcolor{vg}\textbf{AP$_r$}
& \cellcolor{txt}\textbf{AP} & \cellcolor{txt}\textbf{AP$_f$} & \cellcolor{txt}\textbf{AP$_c$} & \cellcolor{txt}\textbf{AP$_r$}
& \cellcolor{mix}\textbf{AP} & \cellcolor{mix}\textbf{AP$_f$} & \cellcolor{mix}\textbf{AP$_c$} & \cellcolor{mix}\textbf{AP$_r$} \\
\hline
A
& \cellcolor{vi}56.3 & \cellcolor{vi}48.4 & \cellcolor{vi}63.7 & \cellcolor{vi}63.8
& \cellcolor{vg}33.6 & \cellcolor{vg}34.8 & \cellcolor{vg}33.4 & \cellcolor{vg}28.0
& \cellcolor{txt}26.5 & \cellcolor{txt}34.5 & \cellcolor{txt}19.7 & \cellcolor{txt}16.0
& \cellcolor{mix}35.2 & \cellcolor{mix}40.0 & \cellcolor{mix}32.3 & \cellcolor{mix}22.8 \\
B
& \cellcolor{vi}57.5 & \cellcolor{vi}49.3 & \cellcolor{vi}64.9 & \cellcolor{vi}66.8
& \cellcolor{vg}35.3 & \cellcolor{vg}36.6 & \cellcolor{vg}35.1 & \cellcolor{vg}29.1
& \cellcolor{txt}29.5 & \cellcolor{txt}34.9 & \cellcolor{txt}25.2 & \cellcolor{txt}20.4
& \cellcolor{mix}36.2 & \cellcolor{mix}38.9{\textcolor{red!70!black}{$\downarrow$}} & \cellcolor{mix}33.8 & \cellcolor{mix}33.1 \\
C
& \cellcolor{vi}63.0 & \cellcolor{vi}54.1 & \cellcolor{vi}70.8 & \cellcolor{vi}73.9
& \cellcolor{hl}33.3{\textcolor{red!70!black}{$\downarrow$}} & \cellcolor{hl}34.7{\textcolor{red!70!black}{$\downarrow$}} & \cellcolor{hl}33.1{\textcolor{red!70!black}{$\downarrow$}} & \cellcolor{hl}26.1{\textcolor{red!70!black}{$\downarrow$}}
& \cellcolor{txt}32.7 & \cellcolor{txt}38.7 & \cellcolor{txt}28.5 & \cellcolor{txt}20.5
& \cellcolor{hl}36.5 & \cellcolor{mix}39.7 & \cellcolor{hl}34.2 & \cellcolor{hl}30.8{\textcolor{red!70!black}{$\downarrow$}} \\
D
& \cellcolor{vi}63.2 & \cellcolor{vi}54.6 & \cellcolor{vi}70.5{\textcolor{red!70!black}{$\downarrow$}} & \cellcolor{vi}74.7
& \cellcolor{hl}33.5 & \cellcolor{vg}35.5 & \cellcolor{vg}32.1{\textcolor{red!70!black}{$\downarrow$}} & \cellcolor{hl}30.0
& \cellcolor{txt}33.2 & \cellcolor{txt}38.9 & \cellcolor{txt}29.4 & \cellcolor{txt}20.3{\textcolor{red!70!black}{$\downarrow$}}
& \cellcolor{hl}38.0 & \cellcolor{mix}40.4 & \cellcolor{hl}36.8 & \cellcolor{mix}31.1 \\
E
& \cellcolor{vi}\textbf{69.2} & \cellcolor{vi}\textbf{61.7} & \cellcolor{vi}\textbf{75.7} & \cellcolor{vi}\textbf{78.3}
& \cellcolor{vg}\textbf{38.6} & \cellcolor{vg}\textbf{39.5} & \cellcolor{vg}\textbf{38.8} & \cellcolor{vg}\textbf{32.8}
& \cellcolor{txt}\textbf{43.0} & \cellcolor{txt}\textbf{45.5} & \cellcolor{txt}\textbf{41.3} & \cellcolor{txt}\textbf{36.6}
& \cellcolor{mix}\textbf{45.2} & \cellcolor{mix}\textbf{47.1} & \cellcolor{mix}\textbf{44.4} & \cellcolor{mix}\textbf{38.7} \\
\hline
\end{tabular}
}
\caption{
Detailed $AP_f$/$AP_c$/$AP_r$ breakdown of component-wise ablation on LVIS-minival across four prompting modes (steps A$\to$E as in Table~\ref{tab:component_wise_ablation}).
}
\label{tab:detailed_ablation_lvis}
\end{table*}

%% file: tables/appendix_efficiency_full.tex
\begin{table*}[t]
\centering
\small
\setlength{\tabcolsep}{3.5pt}
\renewcommand{\arraystretch}{1.1}
\begin{tabular}{clcccccccc}
\hline
\textbf{ID} & \textbf{Evolution Step} & \textbf{Params} & \textbf{Input} & \textbf{GFLOPs} & \textbf{FPS\textsubscript{gen}} & \textbf{Visual-I} & \textbf{Visual-G} & \textbf{Text} & \textbf{Mixed} \\
\hline
A & Baseline (Swin-T) & 115M & 800$\times$1333 & 257 & 15.4 & 56.3 & 33.6 & 26.5 & 35.2 \\
B$_1$ & + DINOv3-ConvNeXt-B & 173M & 800$\times$1333 & 439 & 11.6 & 63.2 & 38.0 & 36.4 & 39.3 \\
B$_2$ & + Reduce to 640$\times$640 & 173M & 640$\times$640 & 208 & 19.4 & 59.1 & 33.4 & 28.6 & 34.4 \\
B$_3$ & + Hybrid encoder & 172M & 640$\times$640 & 164 & 22.0 & 57.5 & 35.3 & 29.5 & 36.2 \\
C & + Prompt-aware dec. & 174M & 640$\times$640 & 165 & 21.3 & 63.0 & 33.3 & 32.7 & 36.5 \\
D & + ICA & 174M & 640$\times$640 & 165 & 21.3 & 63.2 & 33.5 & 33.2 & 38.0 \\
E & + Data engine & 174M & 640$\times$640 & 165 & 21.3 & 69.2 & 38.6 & 43.0 & 45.2 \\
\hline
\end{tabular}
\caption{
Joint efficiency and LVIS-minival accuracy view of the component-wise progression.
IDs match Table~\ref{tab:component_wise_efficiency}.
}
\label{tab:component_wise_efficiency_full}
\end{table*}

%% file: tables/appendix_staged_trex2.tex
\begin{table}[!t]
\centering
\small
\setlength{\tabcolsep}{4pt}
\renewcommand{\arraystretch}{1.1}
\resizebox{\columnwidth}{!}{%
\begin{tabular}{llcccccc}
\hline
\textbf{Model} & \textbf{Training data}
& \multicolumn{3}{c}{\textbf{COCO}}
& \multicolumn{3}{c}{\textbf{LVIS-mv}} \\
\cline{3-8}
& & Visual-I & Visual-G & Text & Visual-I & Visual-G & Text \\
\hline
T-Rex2-T
& O365, OI, HT, CH
& 41.1 & 41.1 & -- & 40.6 & 38.1 & -- \\
& O365, OI, HT, CH, SA-1B
& 56.6 & 38.8 & 45.8 & 59.3 & 37.4 & 42.8 \\
\hline
\end{tabular}%
}
\caption{
Official T-Rex2-T second-stage visual-prompt training results from T-Rex2 Table~6.
Compare with our one-stage progressive scaling in Table~\ref{tab:data_scaling}.
}
\label{tab:data_scaling_trex2}
\end{table}